\documentclass[10pt,journal,compsoc]{IEEEtran}
\ifCLASSOPTIONcompsoc
  \usepackage[nocompress]{cite}
\else
  \usepackage{cite}
\fi
\usepackage[table]{xcolor}

\ifCLASSINFOpdf
\else
\fi
\usepackage{amsmath}
\usepackage{algorithmic}
\usepackage[ruled,vlined]{algorithm2e}
\usepackage{multirow}
\usepackage{amssymb}
\usepackage{color,xcolor}
\usepackage{epstopdf}
\usepackage{graphicx}
\usepackage{makecell}
\usepackage{mathrsfs}
\usepackage{multicol}
\usepackage{multirow}
\usepackage{longtable}
\usepackage{subcaption}
\usepackage{textcomp}
\usepackage{url}
\usepackage{soul}
\soulregister\cite7
\soulregister\ref7
\usepackage{amssymb}
\usepackage{pifont}
\mathchardef\mhyphen="2D

\makeatletter
\newcommand\notsotiny{\@setfontsize\notsotiny\@vipt\@viipt}
\makeatother

\begin{document}
%
\title{MURPHY: Relations Matter \\ in Surgical Workflow Analysis}
%
%
%
%

\author{Shang~Zhao,
        Yanzhe~Liu,
        Qiyuan~Wang,
        Dai~Sun,
        Rong~Liu,
        and~S.~Kevin~Zhou,~\IEEEmembership{Fellow,~IEEE,}
        
\IEEEcompsocitemizethanks{
\IEEEcompsocthanksitem S. Zhao, Q. Wang, D. Sun and S.~Kevin~Zhou are with the School of Biomedical Engineering \& Suzhou Institute for Advanced Research,
Center for Medical Imaging, Robotics, Analytic Computing \& Learning (MIRACLE), 
University of Science and Technology of China, Suzhou, China. Zhou is also affiliated with the Key Lab of Intelligent Information Processing of Chinese Academy of Sciences (CAS), Institute of Computing Technology, CAS, Beijing, China.
\IEEEcompsocthanksitem Y. Liu and R. Liu are with the Faculty of Hepatopancreatobiliary Surgery, the First Medical Center of Chinese PLA General Hospital, Beijing, China.
\IEEEcompsocthanksitem S. Zhao and Y. Liu contributed equally to this work. R. Liu and S. Kevin Zhou (skevinzhou@ustc.edu.cn) are both the corresponding authors.
}
}

%
%

\markboth{Journal of \LaTeX\ Class Files,~Vol.~14, No.~8, August~2015}%
{Shell \MakeLowercase{\textit{et al.}}: Bare Demo of IEEEtran.cls for Computer Society Journals}
%



\IEEEtitleabstractindextext{%
\begin{abstract}
Autonomous robotic surgery has advanced significantly based on analysis of visual and temporal cues in surgical workflow, but relational cues from domain knowledge remain under investigation. Complex relations in surgical annotations can be divided into intra- and inter-relations, both valuable to autonomous systems to comprehend surgical workflows. Intra- and inter-relations describe the relevance of various categories within a particular annotation type and the relevance of different annotation types, respectively. This paper aims to systematically investigate the importance of relational cues in surgery.
First, we contribute the RLLS12M dataset, a large-scale collection of robotic left lateral sectionectomy (RLLS), by curating 50 videos of 50 patients operated by 5 surgeons and annotating a hierarchical workflow, which consists of 3 inter- and 6 intra-relations, 6 steps, 15 tasks, and 38 activities represented as the triplet of 11 instruments, 8 actions, and 16 objects, totaling 2,113,510 video frames and 12,681,060 annotation entities. 
Correspondingly, we propose a multi-relation purification hybrid network (MURPHY), which aptly incorporates novel relation modules to augment the feature representation by purifying relational features using the intra- and inter-relations embodied in annotations. The intra-relation module leverages a Relational Graph Convolutional Network to implant visual features in different graph relations, which are aggregated using a targeted relation purification with affinity information measuring label consistency and feature similarity. The inter-relation module, called Hierarchical Relation Cross Attention, is motivated by attention mechanisms to regularize the influence of relational features based on the hierarchy of annotation types from the domain knowledge. Extensive experimental results on the curated RLLS dataset confirm the effectiveness of our approach, demonstrating that relations matter in surgical workflow analysis.

\end{abstract}

}

\maketitle

\IEEEdisplaynontitleabstractindextext

%
\IEEEpeerreviewmaketitle


\IEEEraisesectionheading{\section{Introduction}\label{sec:introduction}}
\IEEEPARstart{R}{obtic} assisted surgery (RAS), as a raising standard for precise surgical practice, is becoming a new promising platform that facilitates surgical outcomes while improving the operational efficiency~\cite{cleary2004or2020,maier2022surgical}. Through integrated systems and technological advancements~\cite{zhou2019handbook}, especially with the assistance of artificial intelligence (AI), it is possible to make the system to understand the surgical workflows by monitoring and planning the decisions during the procedure with context information, facilitating the surgical outcome and efficiency.

\textbf{Standardizing interpretations of the surgical procedure is necessary for surgical workflow analysis.} The prerequisite of workflow analysis is formulating structural interpretations to disentangle the complex procedural information, and motivating algorithms to precisely extract useful, discriminative information from image contexts. Surgical process modeling (SPM), which is a popular method for surgical education and analysis protocol, has drawn a significant attention from researchers since it defines different categorizations with coarse-to-fine interpretations for surgical procedures. The representative categorization methods include the SPM review, which defines {\it phase, step, activity, and motion}, \cite{lalys2014surgical} and the Society of American Gastrointestinal and Endoscopic Surgeons (SAGES) consensus on surgical video annotation, which defines {\it phase, step, task, and action}~\cite{meireles2021sages}. While these hierarchical interpretations are different, the fundamental requirements are the clear definitions and relationships for hierarchical annotations in order to preserve the consensus and consistency of interpretations. Each interpretation type, including phase, step, activity, motion, task, and action, can be annotated for a surgical video clip. Each type further consists of multiple categories, reflecting the contextual information of visual features, temporal patterns, and relational correlations residing in all annotations.

\begin{figure}[t!]
\includegraphics[width=1.0\linewidth]{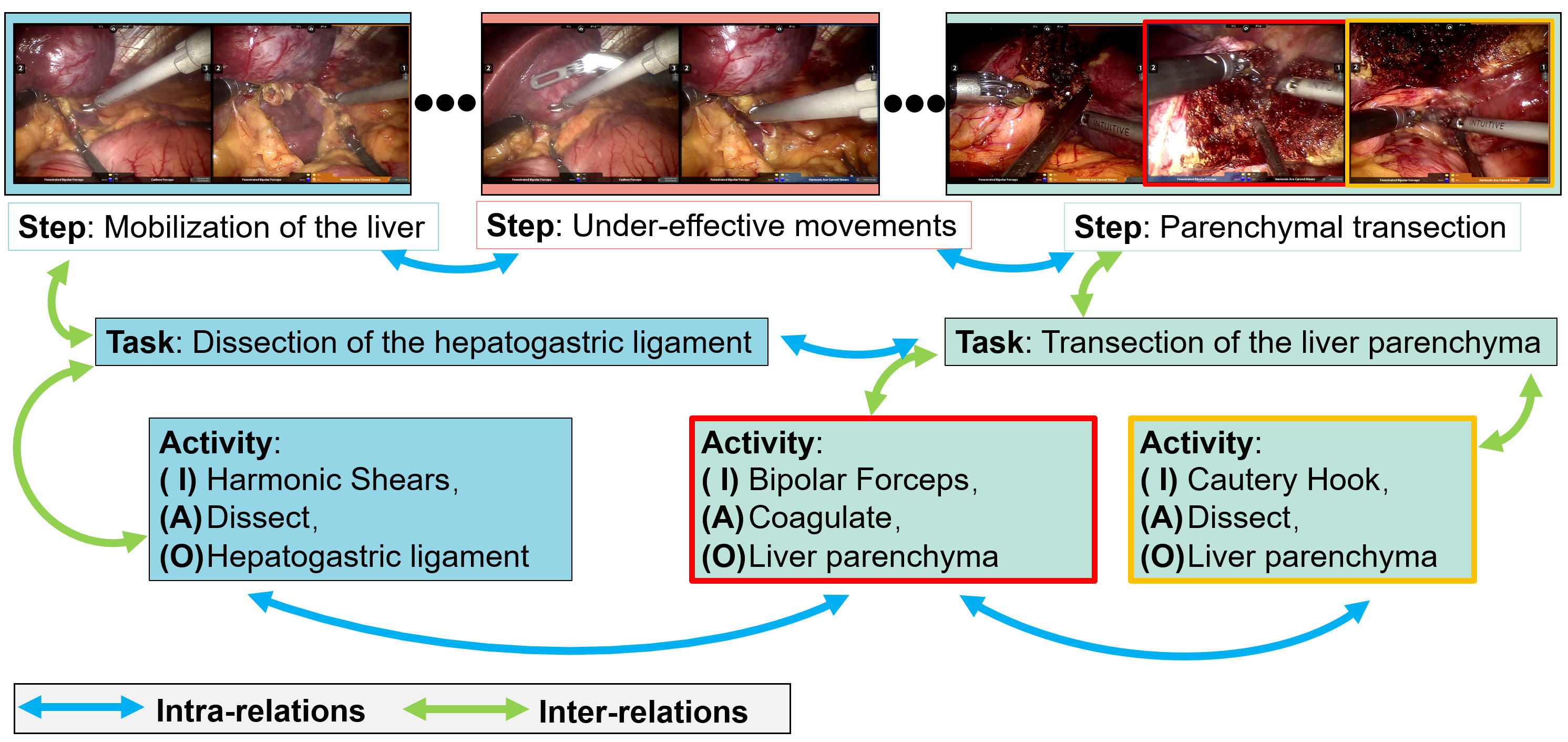}
\caption{The illustration of inter- and intra- relations from a three-level annotation in RLLS12M dataset.} 
\label{fig:inter_intra_relations}
\end{figure}

As the finest-grained event in SPM, surgical activity describes the surgical interaction between instruments and anatomical structures. Recent works~\cite{katic2015lapontospm, nwoye2022rendezvous, nwoye2022cholectriplet2021} have explored the potential of disentangling the activity into an $\langle$IAO$\rangle$ triplet representation, including the instrument (I), the applied action (A), and the targeted object (O). For example, $\langle$cautery hook (I) dissects (A) round ligament (O)$\rangle$. 
The $\langle$IAO$\rangle$ triplet is an expressive form in surgical workflow recognition~\cite{katic2014knowledge}, as it introduces more concrete structural interpretations for activities.
In addition, the recognition of surgical steps and tasks could facilitate the recognition of activity triplets by utilizing the containment information from these hierarchies. Although fine-grained annotation has been discussed in recent years, few works provide systematical analysis across all granularity of hierarchical surgical interpretations and their relations. In this paper, to systematically explore the relations in our hierarchical interpretations, we define {\bf inter- and intra-relations} to respectively describe the containment relationships among different types of annotations and the relations of different categories in the same annotation type.

\textbf{Constructing a hierarchically annotated benchmark dataset is essential to surgical workflow analysis.} Tremendous efforts have been made to investigate utilizing visual and temporal cues for AI-based surgical workflow algorithms. However, it is still under-investigated to leverage relations from surgical domain knowledge in algorithms to facilitate the training outcome. Most existing datasets either focus on one or two types of downstream task, such as phase recognition~\cite{twinanda2016endonet,lea2016surgical,jin2018sv,czempiel2020tecno,jin2021temporal,czempiel2021opera,ban2021aggregating,yuan2022anticipation}, task recognition~\cite{qin2020temporal,van2020multi,jin2020multi,ramesh2021multi}, activity recognition~\cite{gao2020automatic,zhang2021surgical,gao2021trans,nwoye2022rendezvous,van2022gesture}, or have a limited dataset size~\cite{gao2014jhu} for incubating reliable data-driven algorithms. Nevertheless, the lack of large-scale annotated surgical datasets significantly hinders the development of AI-based surgical applications in essential clinical uses~\cite{maier2022surgical}, especially for surgical workflow with multi-grained annotations. 
To this end, \textbf{we contribute the first robotic left lateral sectionectomy (RLLS) dataset, called RLLS12M, with hierarchical annotation}, addressing the issue of lacking benchmark datasets for hierarchical surgical workflow analysis. This dataset consists of 50 RLLS videos of 50 patients operated by 5 surgeons in the Chinese PLA General Hospital, including 2,113,510 annotated video frames with a new hierarchical annotation based on the prior studies~\cite{meireles2021sages, katic2015lapontospm}. Our annotation is a three-level surgical interpretation with 6 annotation types, including 6 steps, 15 tasks, 38 triplet activities, 11 instruments, 8 actions, and 16 objects. All frames are collectively annotated by 3 expert surgeons. There is a total of 12,681,060 annotation entities in the proposed RLLS dataset, making it a unique robotic surgery dataset with hierarchical annotations. With annotated frames, our dataset provides a satisfactory group of relations that includes 3 inter-relations from combinations of annotation types and 6 intra-relations from each annotation type. Fig.~\ref{fig:inter_intra_relations} illustrates the inter- and intra- relations in RLLS dataset. 
For example, there is an inter-relation between the task \textit{transection of the liver parenchyma} and the activity \textit{bipolar forceps coagulate liver parenchyma}, and there is an activity-level intra-relation of \textit{bipolar forceps coagulate liver parenchyma} and \textit{cautery hook dissects liver parenchyma}.

\textbf{Incorporating relational information in hierarchical annotations is beneficial to surgical workflow analysis.} Recent works~\cite{wangsheng2019graph, mao2022imagegcn, jin2022exploring} in medical domain have investigated the potentials of relations in representative tasks such as classification and segmentation tasks, confirming the value of relations to neural network algorithms. Recent works in surgical activity analysis~\cite{nwoye2020recognition, nwoye2022rendezvous, nwoye2022cholectriplet2021} also offer similar observations. Although these works focus on disentangled activity recognition, the discussion is still a single-level annotation recognition without using hierarchical relations. We are unable to completely answer to the question of utilizing the relational information as a general guidance to autonomous surgical workflow analysis without considering hierarchical relations. To the best of our knowledge, few works comprehensively investigate the relation modeling from multi-level annotations. 
In this work, \textbf{we propose a novel multi-relation purification hybrid (MURPHY) network, which targets to improve the feature representation with modules that incorporate inter- and intra-relational information. } 

\begin{table*}[t!]
    \scriptsize
    \setlength\tabcolsep{1.8pt}
    \renewcommand\theadalign{l}
	\renewcommand\cellalign{l}

	\begin{center}
		\caption{Summary of the attributes of existing surgical workflow datasets. There are 8 types of surgeries in the table, which are 
		robot-assisted radical prostatectomy(RRP), 
        proctocolectomy(P), rectal resection(RR), sigmoid resection(SR), laparoscopic cholecystectomy(LC), laparoscopic Roux-En-Y gastric bypass(LRYGB), robotic left lateral sectionectomy(RLLS), and robotic cholecystectomy(RC). We contribute the first RLLS dataset with hierarchical annotations and abundant instrument-anatomy interactions for surgical workflow analysis in this work. The number of annotations indicates the total amount of annotations from all annotated frames. Notably, we also provide under-effective frames, which are unnecessary stylish behaviors.
		}
		\label{tab:related_datasets}
		\begin{tabular}{  l | r | r |  r | l | l | l | r | r |  r | r | r | r | l }
			\hline
			Dataset
			& \#Videos & \#Frames & \#Annotations & Annotation & Anatomy & Operation & \makecell{\#Steps} & \makecell{\#Tasks} & \makecell{\#Activities} & \makecell{\#Instruments} & \makecell{\#Actions} & \makecell{\#Objects} & \makecell{Under-effective \\ frames} \\ 
			\hline
			HeiCo~\cite{maier2021heidelberg} & 30 & $1,499,750$ & $\enspace1,499,750$ &  Step & \makecell{Colon \\ Rectum} &  P, RR, SR & 14 & - & - & - & - & - & No\\
			Cheloc80~\cite{twinanda2016endonet} & 80 & $\enspace\enspace176,110$  & $\enspace\enspace\enspace176,110$ &  Step & \makecell{Cholecyst}   & LC & 7 & - & - & 7 & - & - & No\\
			CholecT50~\cite{nwoye2022rendezvous} & 50 & $\enspace\enspace100,863$ & $\enspace\enspace\enspace403,452$ & Activity & \makecell{Cholecyst}  & LC & - & - & 100 & 6 & 10 & 15 & No\\
			Bypass40~\cite{ramesh2021multi} & 40 & $\enspace\enspace256,000$ & $\enspace\enspace\enspace512,000$ & Step, Task & \makecell{Stomach}  & LRYGB  & 11 & 43 & - & - & - & - & No\\
			PSI-AVA~\cite{Valderrama2022Towards} & 8 & $\enspace\enspace\enspace\enspace73,618$ & $\enspace\enspace\enspace368,090$ & \makecell{Step, Task, \\ Activity} & Prostate & RRP & 11 & 21 & 16 & 7 & 16 & - & No \\
			\hline
			\textbf{RLLS12M (ours)} & 50  & $2,113,510$  & $12,681,060$ & \makecell{Step, Task, \\ Activity} & \makecell{Liver \\ Cholecyst}  & RLLS, RC  & 6 & 15 & 38 & 11 & 8 & 16 & Yes\\

			    \hline
		\end{tabular}
	\end{center}
\end{table*}

Our contributions are as follows:
\begin{itemize}
    \item We contribute the first RLLS surgery dataset with millions of hierarchical annotation entities for surgical workflow analysis, providing a new large-scale benchmark and advocating the surgical workflow analysis with hierarchical interpretations. 
    \item We propose a new R-GCN relation purification module to enhance the feature representation with relational information embedded in annotations. The R-GCN module aggregates relational features based on inter- and intra- relations in annotations, encouraging aggregation with high label consistency and backbone feature similarity. 
    \item We propose a new hierarchical relation cross attention (HRCA) module to extract  inter-relations from the embeddings in multi-grained annotations, regularizing the classification result with domain knowledge.  
\end{itemize}

\section{Related works}\label{sec:related_works}
\subsection{Public Surgical Datasets}
Surgical workflow analysis ideally requires multiple annotation types to comprehensively describe the surgical state at each sampling time and therefore, existing datasets~\cite{twinanda2016endonet,aksamentov2017deep,ahmidi2017dataset,leibetseder2020glenda,maier2021heidelberg,ramesh2021multi,nwoye2022rendezvous} usually contain one or more types of annotations. Although names of annotation types can vary among different datasets, consensus on annotation types can be generally categorized as: \textit{step} is a temporal phase that describe the stage of the progression in a high-level statement; \textit{task} is a fundamental surgical event in a relative smaller temporal window that aims to accomplish a combination of surgical activities; \textit{activity} is a surgical action to perform concrete movements that can be named with a precise description of instrument-tissue interactions.

Current public datasets about surgical workflow analysis are mainly in the form of videos and images, as shown in Table \ref{tab:related_datasets}. Laparoscopic cholecystectomy (LC) is the most common procedure for surgical datasets because it is the generally accepted gold standard surgery~\cite{garrow2021machine}. In addition, the majority of datasets, among which the popular ones are HeiCo~\cite{maier2021heidelberg} and Cheloc80~\cite{twinanda2016endonet}, only provide phase or step annotations, resulting in limited granularity for algorithms to understand the surgery progression~\cite{aksamentov2017deep, leibetseder2020glenda}. Recently, ChelocT50~\cite{nwoye2022rendezvous} is available for the investigation of the semantic relations inside activity descriptions, which allows AI to understand the component-level relations in endoscopic views. However, it is challenging to make the algorithms to recognize abundant activities with only the finest-grained information due to the lack of a big picture. The most intuitive thought would be integrating multiple-level annotations to make AI progressively understand the procedure. Bypass40~\cite{ramesh2021multi} coincides with this thought and presents an LRYGB dataset that consists of step and task annotations. The lack of triplet activity annotations hinders us from comprehensively investigating surgical workflow analysis with a multi-level coarse-to-fine design. The PSI-AVA dataset~\cite{Valderrama2022Towards} provides multi-leveled annotations with varying granularities, but it emphasizes the instrument descriptions while lacking the annotations of interacting objects. To this end, we contribute the first RLLS dataset that contains all the aforementioned annotations, along with under-effective frames that have not been investigated in this domain yet. The dataset includes different anatomies and types of surgery with a relatively large population of patient videos, bridging the gap of lacking a hierarchical clinical dataset.

\subsection{Relations in Surgical Workflow Recognition}
Abundant relations in the surgical interpretation embody natural constraints for designing robust analysis algorithms. For example, the step of the procedure can be represented as a directional graph that allows cycles because the surgeon may need to revisit the step to polish the objective until reaching a satisfactory outcome. Benefiting from the strict definitions of the annotation hierarchy, we can easily partition the tasks and activities with respect to the step category. We denote the relations of categories within the same annotation type as intra-relations. Correspondingly, the relations among different annotation types are named as inter-relations.

The majority of existing works prefer applying an encoder that utilizes spatial and temporal salient patterns for surgical workflow analysis. However, analyzing surgical procedures with miscellaneous relations embedded in annotations has not been discussed thoroughly. Although relations in temporal space, such as Recurrent Neural Network-based methods~\cite{dipietro2016recognizing, kannan2019future, van2020multi, jin2021temporal, ramesh2021multi}, have been investigated in recent workflow analysis methods, none of these works attempts to systematically explore the relations across different categories within a specific annotation type. Notably, recent works~\cite{nwoye2020recognition, nwoye2022rendezvous, long2021relational} also put efforts into analyzing the relations in the triplet activity representation. The component-based activity provides finer interpretations for describing a surgeon's behavior, enabling the network to comprehend the activity through weakly-supervised component information by using class activation maps. However, this family of methods only considers the activity-level descriptions without incurring the hierarchy of annotations, resulting in the lack of incorporating multi-level relational patterns. Recently, the medical imaging community favors investigating the possibility of extracting discriminative relational patterns~\cite{wangsheng2019graph, mao2022imagegcn, ding2022graph} that are different from the approaches that use spatial and temporal features. Unlike existing workflow analysis methods, we take a different strategy that utilizes the GCN to learn the relations of categories in each annotation type. Besides, we also model the constraints in hierarchical annotations to comprehensively regularize the framework with both inter- and intra-relations, contributing a new approach to embedding relations in network architectures.

\begin{table}[t!]
    \scriptsize
	\setlength\tabcolsep{3.5pt}
	\renewcommand\theadalign{l}
	\renewcommand\cellalign{l}

	\begin{center}
		\caption{Summary of the existing works for surgical procedure analysis. Different from existing representative works, we exploit hierarchical relations in three-level annotations for feature representation learning. }
		\label{tab:related_methods}
		\begin{tabular}{  l | l | l | l   }
			\hline
			Method
			& 	Primary task & 	Relation type & 	Approach  \\ \hline
			AttentionTriplet~\cite{nwoye2020recognition} & Activity & \makecell{Triplet \\ components} & Class activation \\
			Rendezvous~\cite{nwoye2022rendezvous} & Activity & \makecell{Triplet \\ components} & \makecell{Attention, \\ Class activation} \\
			MTMS-TCN~\cite{ramesh2021multi} & Step, Task & Temporal & Temporal convolution  \\
			\hline
			\textbf{MURPHY (ours)} & \textbf{\makecell{Step, Task, \\ Activity}}  & \textbf{Hierarchical} & \textbf{R-GCN, HRCA} \\

			\hline
		\end{tabular}
	\end{center}
\end{table}

\subsection{Graph Convolutional Networks for Classification }
Graph Convolutional Networks (GCN)~\cite{kipf2017semi} define the convolution operation on graphs by a differentiable message-passing mechanism that aggregates informative representations from nodes and edges, improving the classification performance in complex tasks. Much work has been done to explore the graph-based image feature extraction, graph-based label correlation analysis, and graph-based hyper-pixels for spectral image classification~\cite{chen2022dynamic}. This convolution scheme motivates the investigation on addressing multiple relation modeling problems with GCN. Relational Graph Convolutional Networks (R-GCN)~\cite{schlichtkrull2018modeling}, as a representative network, formulate varying types of relations into a graph structure and define a relational message-passing propagation model for link prediction and node classification. ImageGCN~\cite{mao2022imagegcn} proposes an R-GCN variant for chest X-ray classification, which models the demographic information as relations that compose a graph for X-ray scans. However, it models the affinity information from all annotated labels in the data, which introduces side-effects in the following two scenarios: 1) the quantity of annotation is extremely large which leads to an unfriendly computation efforts; 2) limited annotations or the data collection can hardly cover the original distribution so that the pre-computed affinity matrix poorly reflects true correlations among labels. To avoid tedious computation and reliance on labels for affinity information, we propose a feature similarity-based R-GCN framework for surgical activity analysis along with a batch-wise affinity learning method to model the label correlations. Tool presence~\cite{wangsheng2019graph} has been discussed in the surgical analysis domain with GCN, but the relation-embedded methods still remain under investigation. In this work, we conduct comprehensive investigations to model relations in surgical procedures with both graph convolutions and attention mechanisms.

\begin{figure*}[t]
\centerline{\includegraphics[width=\linewidth]{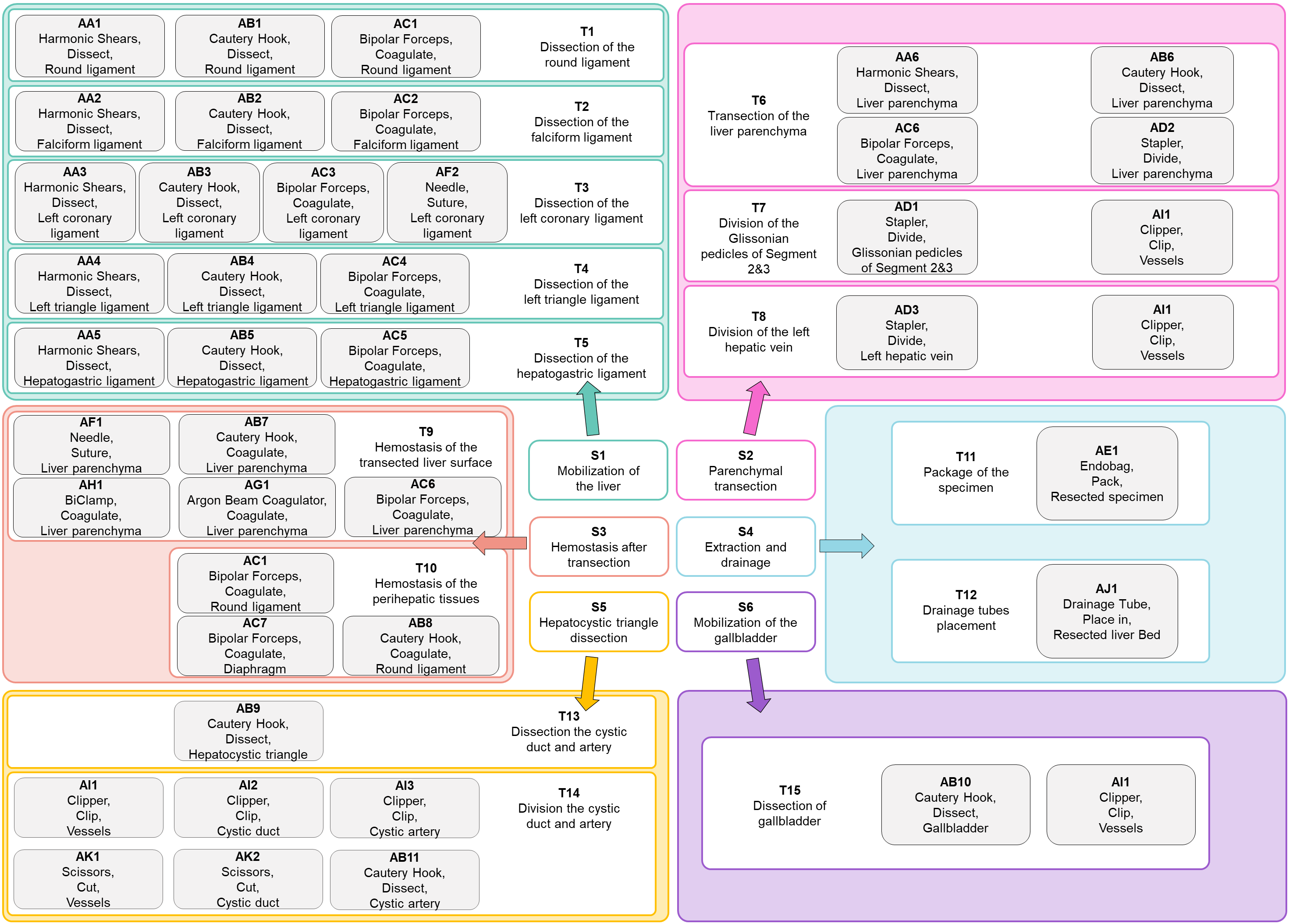}}
\caption{Summary of the ST$\langle$IAO$\rangle$ hierarchy of temporal annotations in the RLLS12M dataset. The blobs with the same color indicate that they belong to the same sub-hierarchy. Note that activity annotation is the fundamental element in our surgery interpretation, which has multiple occurrences across particular groups of tasks, such as \textbf{AI1}.}
\label{fig:rlls_label_summary}
\end{figure*}

\section{The RLLS12M Dataset}

\subsection{Surgical Data Collection}

As of now, the majority of public videos of robotic surgery is based on the in-vitro and ex-vivo experiments \cite{ahmidi2017dataset, sarikaya2017detection}. Our RLLS12M datasets contain 50 in-vivo videos in total from 50 patients (one video per patient), including 46 regular RLLS cases and 4 cases of RLLS combined with robotic cholecystectomy. All videos are retrospectively acquired from the Faculty of Hepato-pancreato-biliary Surgery, Chinese PLA General Hospital since Mar 2017 to Mar 2020. All collected surgeries are operated by 5 expert surgeons, using the da Vinci Si System (Intuitive Surgical Inc., Sunnyvale, CA, USA), so that the variety of surgery styles is preserved. Different from other types of surgery with plenty of interfering factors, RLLS operations generally contain intuitively cogent scenarios, making them appropriate for modeling. Each surgical video contains the entire clinical procedure, which also records abundant challenging scenarios during surgery, such as smoke and high specularity. All videos are captured at 60 fps, 1280×1024 pixels from the robotic camera with a 30-degree oblique and magnification of up to 10-12x. To provide a friendly experimental configuration and storage efficiency, we decrease the resolution to 640×512 pixels and down-sample the frame rate to 25 fps.

\subsection{Hierarchical Surgical Annotation}
\label{sec:RLLS_annotation_description}
We apply a hierarchical surgical interpretation, denoted as ``ST$\langle$IAO$\rangle$", in our RLLS12M dataset with 6 annotation types according to the consensus in SAGES~\cite{meireles2021sages}, including step (S), task (T), triplet activity ($\langle$IAO$\rangle$), instrument (I), action (A), and object (O) that are listed in a coarse-to-fine order. Notably, a step contains multiple tasks, and a task contains multiple triplet activities, and the I, A, O component-level annotations are located at the same level in the hierarchical annotation. This ST$\langle$IAO$\rangle$ surgical interpretation is discussed by a panel of senior experts in the Chinese PLA General Hospital considering both previous reports of multiple medical centers and our experiences~\cite{hu2019robotic, zhu2022comparison}. The summary of our hierarchical RLLS interpretation is shown in Fig.~\ref{fig:rlls_label_summary}. In detail, step represents the specific parts of surgery to accomplish a clinically meaningful goal, such as mobilization of the liver, which preserves both the precise objective and the abstraction of content. To provide more accurate descriptions, we define the task as a lower-level interpretation to represent a more concrete temporal event for a smaller goal. 
For instance, mobilization of the liver requires accomplishing the tasks of dissecting the round ligament and the falciform ligament. Activity denotes the fundamental behavior for the surgeon to conduct a necessary operation, which is the finest element in our interpretation. We use the $\langle$IAO$\rangle$ triplet activity form to represent the instrument used, the action performed, and the object acted upon \cite{katic2014knowledge}. For example, the harmonic shears dissect the round ligament expressed as $\langle$harmonic shears (I) dissect (A) round ligament (O)$\rangle$. Note that all aforementioned annotations except the component annotations can be interrupted and repeated without following a specific order in the practical clinical environment. We treat the step, task, and activity as the primary annotations that offer the situational explanations in varying granularities, and treat the other 3 component-level interpretations as auxiliary annotations for facilitating the understanding of concrete interactions in the surgery. 
In addition, we also define under-effective frames~\cite{meireles2021sages} for those movements or events that have less necessity in the procedure. Since different experts have their own preferences for performing the surgery, it is inevitable to bring in some subjective stylish movements during the procedure. Moreover, the procedure may exist in some idle temporal segments where the team may be discussing a situation or waiting for the examination results. These frames are also annotated so that it provides the model with the capability of distinguishing under-effective events in the surgery.

All frames in this dataset are labeled with the proposed ST$\langle$IAO$\rangle$ hierarchical surgical annotation by 3 expert surgeons who have more than 5-year clinical experience. Due to the complexity of annotation designs, we design a strict protocol for labeling frames to minimize the interference of subjective factors in annotated frames. First, the hierarchical annotations should have precise time alignment. For example, when we define the temporal range of a step, we treat the first frame of the first task and the last frame of the last task in this step as the beginning and the ending frame for the step. The same principle applies to the task and activity as well. In this work, we aim to validate the effectiveness of hierarchical surgical interpretations so that the definitions of activity are only for the 11 main instruments.
We also define a clear rule for identifying activity: the activity starts when the instrument is effectively acting on the object and ends when the activity is complete. For an under-effective event, we annotate it when the time interval between two effective activities exceeds 5 seconds. Expert raters can minimize their subjective variance of annotation by following these principles. The concordance correlation coefficients for all 3 rater pairs are 0.82, 0.89, and 0.84, confirming their agreements on annotation.

In sum, we contribute the RLLS12M dataset, a large-scale collection of images and annotations of RLLS, curating 50 videos of 50 patients operated by 5 surgeons and annotating a hierarchical workflow, which consists of 6 steps, 15 tasks, and 38 activities represented as the triplet of 11 instruments, 8 actions, and 16 objects and forms 3 inter- and 6 intra-relations. It contains a total of 2,113,510 video frames and 12,681,060 annotation entities. Refer to \url{http://miraclelab.site/?page_id=2224} to access the RLLS12M dataset.

\section{Methods}
\subsection{Overview}
\begin{figure*}[t]
\centerline{\includegraphics[width=\linewidth]{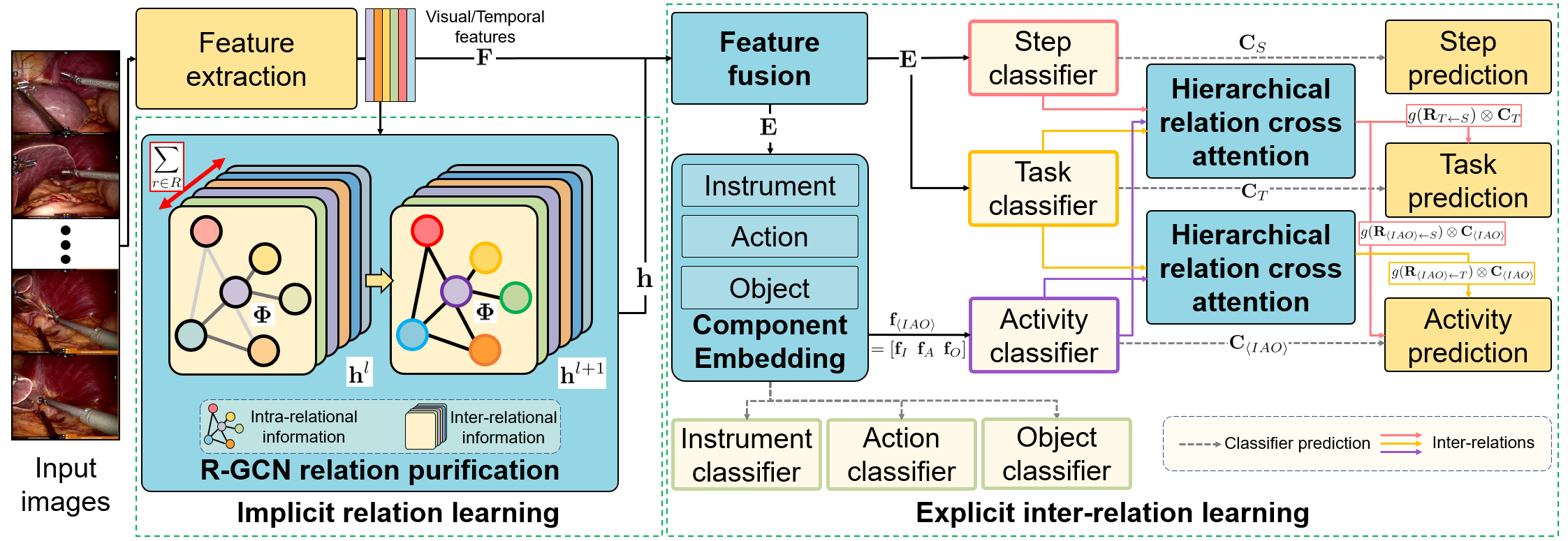}}
\caption{The architecture of multi-relation purification hybrid (MURPHY) network.}
\label{fig:framework}
\end{figure*}

The proposed MURPHY framework for surgical analysis, as shown in Fig.~\ref{fig:framework}, contains the following modules: 1) image feature extraction module; 2) implicit intra- and inter-relation learning module; 3) explicit inter-relation learning module. The input data are surgical images and the outputs are recognition of surgical step, task, and triplet-based activity as {\it primary tasks} and recognition of instrument, action, and object as {\it auxiliary tasks}. The feature extraction module utilizes the Convolutional Neural Networks (CNN) and Long Short Term Memory (LSTM) model to generate informative representations. Unlike existing methods that only utilize label distributions and component interactions~\cite{nwoye2020recognition, nwoye2022rendezvous}, we take advantage of GCN and the attention mechanism to present new modules to learn relations in the surgery procedure. Detailed explanations on relation modeling will be introduced in the sections \ref{sec:rgcn_model} and \ref{sec:hrca}.

\subsection{Feature Extraction}\label{sec:feature_extraction}
The proposed framework first translates the input images to descriptive visual feature representations by using a feature extraction network. This deep visual feature can better describe the content information for relation learning. In our experiment, we extract visual features with the CNN and LSTM to demonstrate the robustness of visual feature for our work. The backbone module takes image frames as inputs and extracts the corresponding backbone feature $\mathbf{F} \in \mathbb{R}^{B\times D}$, where $B$ is batch size and $D$ is the backbone feature dimension. 

\subsection{Implicit Relation Learning with R-GCN}\label{sec:rgcn_model}

\textbf{We apply a modified R-GCN module to implicitly learn intra- and inter-relations.} 
We construct a graph $\mathbf{G}$ based on the input features in each mini-batch, where each node is $\mathbf{F}_{i} \in \mathbb{R}^{D}; i\in \{1,2,...,B\}$. Instead of applying a pre-defined adjacency matrix, we dynamically construct the adjacency matrix $\mathbf{S}^{r} \in \mathbb{R}^{B\times B}$ to represent the edge connectivity for all nodes under a certain intra-relation $r \in R$ in the batch, where $R$ is the total number of annotation types. 
Towards utilizing the natural relations among the different classes in the surgical procedure, relation modeling not only aims to comprehensively aggregate the useful information from inter-relations in annotation types, but also individually evaluates the similarity between different categorical labels to quantify valid relational information for ST$\langle$IAO$\rangle$ recognition. 
To this end, we present an R-GCN module for learning relations among categorical labels at the same label granularity and aggregating hierarchical relational information from different annotations.

\subsubsection{Dynamic Similarity Matrix Construction} \label{sec:dynamic_matrix}
\textbf{The adjacency construction for GCN is critical, we propose a dynamic building strategy.} The adjacency matrix in the graph convolution can not only describe the node connectivity but also the status quo of feature correlations in nodes~\cite{mao2022imagegcn}. Existing R-GCN designs in the medical analysis domain require an expensive computational cost of adjacency matrix that takes the entire dataset into account to build the global structural information, which is too expensive for surgical scene analysis because each surgical video generally contains more than thousands of frames as graph nodes. Inspired by dynamic adjacency construction for GCN~\cite{wangsheng2019graph}, we extend this approach in our R-GCN module with a hybrid design that integrates both label consistency and feature correlation in frames. The final adjacency for relation $\mathbf{S}^{r}$ is the Hadamard product of feature correlation tensor $\mathbf{M}$ and label consistency tensor $\mathbf{A}^{r}$ for a specific relation, as shown below:
\begin{equation}
\mathbf{S}^{r} = \mathbf{M} \otimes \mathbf{A}^{r},
\label{eq:g_matrix}
\end{equation}
where the feature correlation tensor $\mathbf{M}$ is constructed with the normalized cosine similarity to measure correlations of backbone feature $\mathbf{F}$ in the batch,
\begin{equation}
\begin{aligned}
&\mathbf{M}_{i, j} =  \frac{\exp(s_{ij})}{\sum_{k=1}^{B} \exp(s_{ik})}, 
&\text{where } s_{i,j} = \frac{\left(\mathbf{F}_{i} \cdot \mathbf{F}_{j}\right)}{\parallel\mathbf{F}_{i}\parallel_{2} \cdot \parallel\mathbf{F}_{j}\parallel_{2}}. 
\end{aligned}
\label{eq:feature_corr_matrix}
\end{equation}
The label consistency tensor $\mathbf{A}^{r}$ is an indicator matrix that flags the pairs of samples belonging to the same class,
\begin{equation}
\mathbf{A}^{r}_{i, j} = \left\{
\begin{aligned}
&1, & \text{if } \mathbf{y}_{i}^{r} = \mathbf{y}_{j}^{r}, \\
&0, & \text{if } \mathbf{y}_{i}^{r} \neq \mathbf{y}_{j}^{r},
\end{aligned}
\right.
\label{eq:label_consistency_matrix}
\end{equation}
where $\mathbf{y}_{i}^{r}$ is the label category under relation $r$. In this way, we can assemble the domain knowledge into the relation learning module, resulting in a configurable framework. The label consistency provides a precise indication to guide relational feature learning. The combination of the two tensors describes the adjacency information by similarity measured from different aspects, which only encourages adjacency belonging to the same class during training. There is no label consistency information during inference thus we only keep the feature correlation part for inference.  


\subsubsection{Relation Purification}
\textbf{We utilize the massage passing scheme in R-GCN to purify the relational feature through R-GCN layers.} In GCN, node hidden features are evaluated by aggregating both the node and structure information in the graph.
The graph convolution fuses each node feature with its neighboring node features by using an adjacency matrix, resulting in the graph convolutional feature with considering the local structural information~\cite{kipf2017semi}:

\begin{equation}
\mathbf{h}_{i}^{l+1} = \varphi \left( \sum_{k \in N(i)} \mathbf{\hat{A}}_{i,k}\mathbf{h}_{k}^{l}\mathbf{W}^{l}  \right),
\label{eq:gcn}
\end{equation}
where $\mathbf{h}_{i}^{l+1} \in \mathbb{R}^{H}$ is the hidden feature for the next layer $l+1$, and $H$ is the feature hidden dimension. $\mathbf{h}_{i}^{l}$ is the backbone feature $\mathbf{F}_{i}$ when $l=0$. $\mathbf{h}_{k}^{l}$ denotes the hidden state of neighboring node $k \in N(i)$ in the current layer $l$. $\varphi$ is a non-linear activation function. The matrix $\mathbf{\hat{A}}$ is generally represented by a symmetrically normalized graph Laplacian matrix, where $\mathbf{\hat{A}}_{i,k}$ is the adjacency between node $i$ and $k$. 


\textbf{We derive a relation learning module that implicitly learn the intra- and inter-relational information through relational graph convolution.} The R-GCN module purifies the meaningful information from the structural views of varying intra-relations by convolution. In each convolutional layer, it accumulates the intra-relational information across all types of relations to an aggregated hidden feature as the output. To learn the informative feature from intra-relations $R$, we adapt the Eq.~(\ref{eq:gcn}) to define the aggregation function $\mathbf{\Phi}$ that accumulates the local neighbor information under a certain annotation type  $r \in R$,
\begin{equation}
\begin{aligned}
\mathbf{\Phi}\left( \mathbf{S}^{r},  \mathbf{h}^{l}, i\right) = \sum_{k \in N_{r}(i)} \mathbf{S}_{i,k}^{r} \mathbf{h}_{k}^{l} \mathbf{W}_{r}^{l} ,
\end{aligned}
\label{eq:intra_accum}
\end{equation}
where $N_{r}(i)$ is the neighbor set of the current node $i$ under relation $r$. $\mathbf{S}_{i,k}^{r}$ describes the similarity between the neighboring node $k$ and the current node $i$.  
$\mathbf{W}_{r}^{l}$ is the trainable parameters for the intra-relation $r$ in the layer $l$. $\mathbf{h}_{k}^{l}$ is the hidden feature in the neighbor set. Furthermore, we apply the relational graph convolution based on Eq.~(\ref{eq:intra_accum}), which aggregates intra-relational features across annotation types to realize the inter-relational information accumulation:
\begin{equation}
\begin{aligned}
\mathbf{h}_{i}^{l+1} 
&= \varphi \left( \sum_{r \in R} \mathbf{\Phi}\left( \mathbf{S}^{r},  \mathbf{h}^{l}, i\right) \right) \\
&= \varphi \left( \sum_{r \in R} \sum_{k \in N_{r}(i)} \mathbf{S}_{i,k}^{r} {\mathbf{h}_{k}}^{l} \mathbf{W}_{r}^{l}  \right),
\end{aligned}
\end{equation}
The hidden state implicitly includes the interactions of different relational features, influencing the R-GCN to learn from aggregated inter-relations. With the accumulation across inter-relations, the relational graph convolution is finalized with the following representation:
\begin{equation}
\mathbf{h}_{i}^{l+1} = \varphi \left( \sum_{r \in R} \sum_{k \in N_{r}(i)} \mathbf{S}_{i,k}^{r} {\mathbf{h}_{k}}^{l} \mathbf{W}_{r}^{l}  + \mathbf{h}_{i}^{l} \mathbf{W}_{0}^{l} \right).
\end{equation}
We follow the original design~\cite{schlichtkrull2018modeling} to introduce an $\mathbf{h}_{i}^{l}\mathbf{W}_{0}^{l}$ term that incorporates self-connection of a special relation type for each node $i$. $\varphi$ is the ReLU activation function.
Specifically, we apply a layer normalization at the beginning of the R-GCN block, aiming to avoid interference on features due to the scale effects in parameters. 
Once the relational feature is produced by our R-GCN module, we fuse the relational feature with backbone features for prediction. The step and task classifiers take the fused feature $\mathbf{E} = [\mathbf{F}\enspace \mathbf{h}] \in \mathbb{R}^{B \times (D+H)}$ by concatenating the backbone feature and relational feature for final prediction.

\subsection{Explicit Inter-Relation Learning with HRCA}\label{sec:hrca}
\subsubsection{Component Embedding for HRCA}
\textbf{We derive a component embedding block for HRCA by explicitly learning the inter-relations among I, A, O with supervision. }
Component-level information in I, A, O has important inter-relations that correlate with $\langle$IAO$\rangle$. Furthermore, component feature embedding integrates new discriminative information to enrich the representation power of the fused feature. Inspired by the previous works~\cite{nwoye2022rendezvous}, we put activity recognition into a special design that
deliberately projects the fused feature $\mathbf{E}$ onto each triplet component $\mathbf{f}_{c}$ in the activity label, aiming to strengthen the distinguishing power of component features. 
\begin{equation}
\mathbf{f}_{c} = \varphi \left(\mathbf{\Gamma}_{c} \mathbf{E} + \mathbf{b}_{c} \right),
\label{eq:triplet_embed}
\end{equation}
where $c \in \{$I, A, O$\}$, $\mathbf{\Gamma}_{c}$ and $\mathbf{b}_{c}$ are learnable affine transformation parameters.
These components are supervised by the corresponding categorical labels with individual classifiers so that we can explicitly disentangle the component-level information through the non-linear projection. Moreover, the component embedding makes the HRCA-integrated classifier consider the inter-relations among the three types of annotations. All perspective relational features are assembled to jointly predict the triplet activity and therefore, the special embedding for activity is a concatenated tensor, $\mathbf{f}_{\langle IAO\rangle} = [\mathbf{f}_{I} \enspace \mathbf{f}_{A} \enspace \mathbf{f}_{O}].$

\subsubsection{HRCA}
\textbf{Benefiting from the component embedding, we further build HRCA for deliberately constraining the inter-relations for learning with hierarchical annotations.} 
The containment relationships in hierarchical annotations can be concluded that each step comprises several tasks and each task contains multiple activities. 
We realize this scheme with a cross-attention mechanism named the HRCA module.
HRCA is a special form of self-attention based on a pre-defined domain knowledge matrix, where the knowledge matrix provides the affinity information among categorical labels in two different types of annotations. For example, a step "\textit{parenchymal transection}" consists of tasks such as "\textit{transection of the liver parenchyma}", "\textit{division of the Glissonian pedicles of Segment 2\&3}", and "\textit{division of the left hepatic vein}". The higher confidence of prediction on step results in stronger constraints in predicting the task from the three potential options. In addition, this pre-defined domain knowledge scheme also has robustness since it can be customized based on the specific surgical procedure.

To formulate these insights, our HRCA module first applies an embedding layer with one-dimensional convolutions to project the classifier outputs $\mathbf{C}_{i}\in\mathbb{R}^{B \times 1 \times T_{i}} $ into hidden space $\mathbf{Q}_{i}\in\mathbb{R}^{B \times K \times T_{i}}$ for correlation evaluation, where $T_{i}$ denotes the number of categories of the annotation type. Correlation is evaluated pair-wisely between different types of features $\mathbf{Q}_{i}$ and $\mathbf{Q}_{j}$, where $i\neq j$. The cross-attention is the product between the correlation tensor with softmax normalization and the pre-defined knowledge tensor $\mathbf{M}_{i,j}^{\text{prior}}\in\mathbb{R}^{B\times T_{j} \times T_{i}}$. The knowledge tensor is a binary tensor in which the non-zero entries indicate containment relationships between corresponding pairs of categorical labels in different annotations. The final attention result $\mathbf{R}_{i\leftarrow j}\in\mathbb{R}^{B \times T_{j} \times T_{i}}$ from annotation type $j$ to $i$:

\begin{equation}
\mathbf{R}_{i\leftarrow j} = \text{softmax}(\frac{\mathbf{Q}_{j}^{\top} * \mathbf{Q}_{i}}{\sqrt{d_i}} )\otimes \mathbf{M}_{i,j}^{\text{prior}} * \mathbf{Q}_{i}^{\top}, 
\label{eq:hrca1}
\end{equation}
where $B$ denotes the batch size.
Note that the transpose operation does not affect the batch dimension. The attention feature then influences the corresponding target feature on the original classifier output as the final prediction:
\begin{equation}
\mathbf{\hat{C}}_{i} = \mathbf{C}_{i} + g(\mathbf{R}_{i\leftarrow j}) \otimes \mathbf{C}_{j} + g(\mathbf{R}_{i\leftarrow k}) \otimes \mathbf{C}_{k}, 
\label{eq:hrca_assemble}
\end{equation}
where $i,j,k \in \{S, T, \langle IAO\rangle\}$ and $i\neq j\neq k$, $g$ represents the global average pooling operator for the hidden dimension.

\begin{figure}[t!]
\centerline{\includegraphics[width=\linewidth]{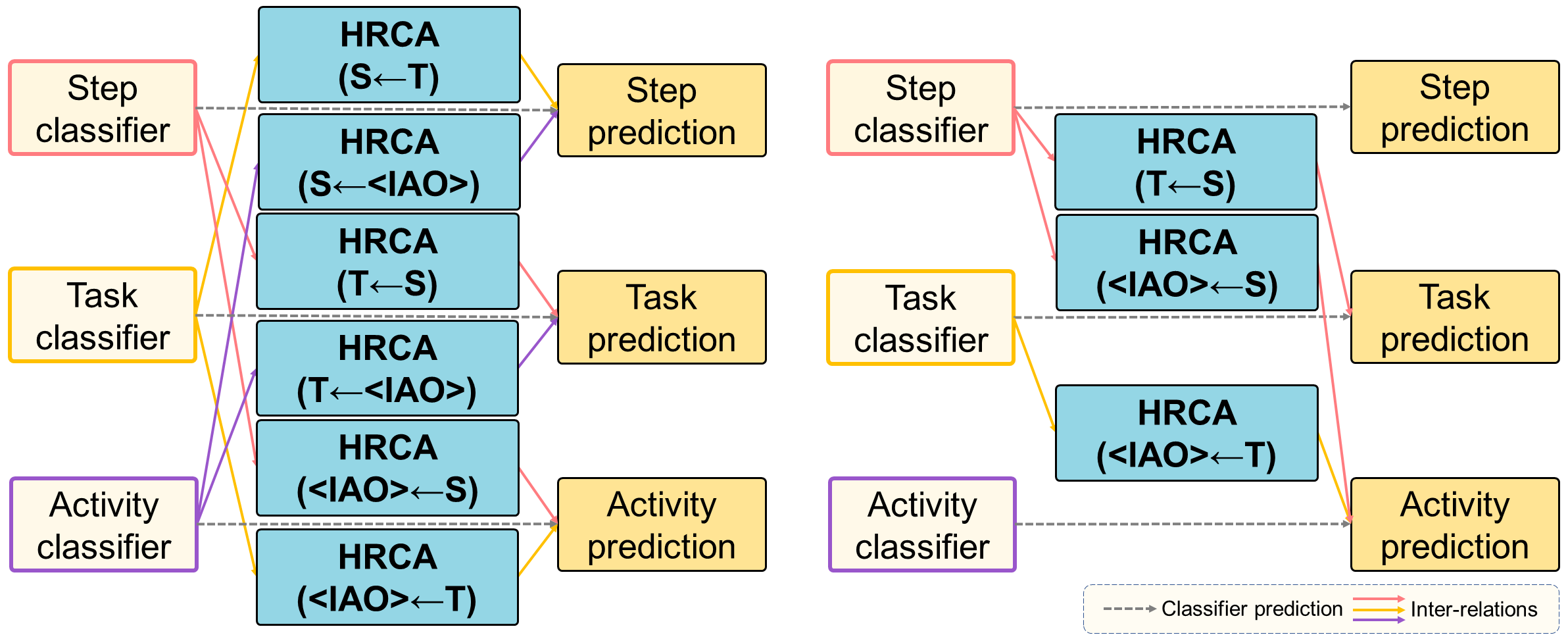}}
\caption{Two HRCA aggregation schemes, including the full combination version (left) and the coarse-to-fine version (right). We ignore arrow lines for the $\mathbf{Q}_{i}^{\top}$ part in Eq.~(\ref{eq:hrca1}) for clear illustration.}
\label{fig:hrca_agg}
\end{figure}

\subsubsection{Coarse-to-fine HRCA Aggregation Scheme}
\textbf{We apply a coarse-to-fine scheme to avoid interference from noisy representations in finer-leveled features.} The integrating scheme for HRCA is an important factor in the prediction results. The intuitive solution is evaluating all combinations among three annotation types in the RLLS12M dataset shown in Fig.~\ref{fig:hrca_agg}, which means the network evaluates 6 pair-wise cross-attention tensors to augment each type of final prediction by the other two. 
The smaller number of categorical results is prone to empirically making the networks easier to make more confident decisions. Inspired by this, we continue to conduct varying combinations for HRCA. This observation brings the idea of a coarse-to-fine HRCA aggregation scheme, which is augmenting the finer-grained annotation recognition with the cross-attentions from the coarser annotations, with the following formulation:
\begin{equation}
\begin{aligned}
\mathbf{\hat{C}}_{T}&=\mathbf{C}_{T} + g(\mathbf{R}_{T\leftarrow S}) \otimes \mathbf{C}_{T}, & \\
\mathbf{\hat{C}}_{\langle IAO\rangle}&=\mathbf{C}_{\langle IAO\rangle} + g(\mathbf{R}_{\langle IAO\rangle\leftarrow T})\otimes\mathbf{C}_{\langle IAO\rangle} & \\ 
& + g(\mathbf{R}_{\langle IAO\rangle\leftarrow S})\otimes\mathbf{C}_{\langle IAO\rangle}. &
\end{aligned}
\label{eq:hrca_c2f_assemble}
\end{equation}
Therefore, this coarse-to-fine manner potentially makes the coarse-level tasks to avoid interference from the relatively difficult recognition tasks ({\it e.g.}, activity recognition w.r.t task recognition), which results in performance degradation.

\begin{figure}[t]
\centerline{
\includegraphics[width=0.226\linewidth]{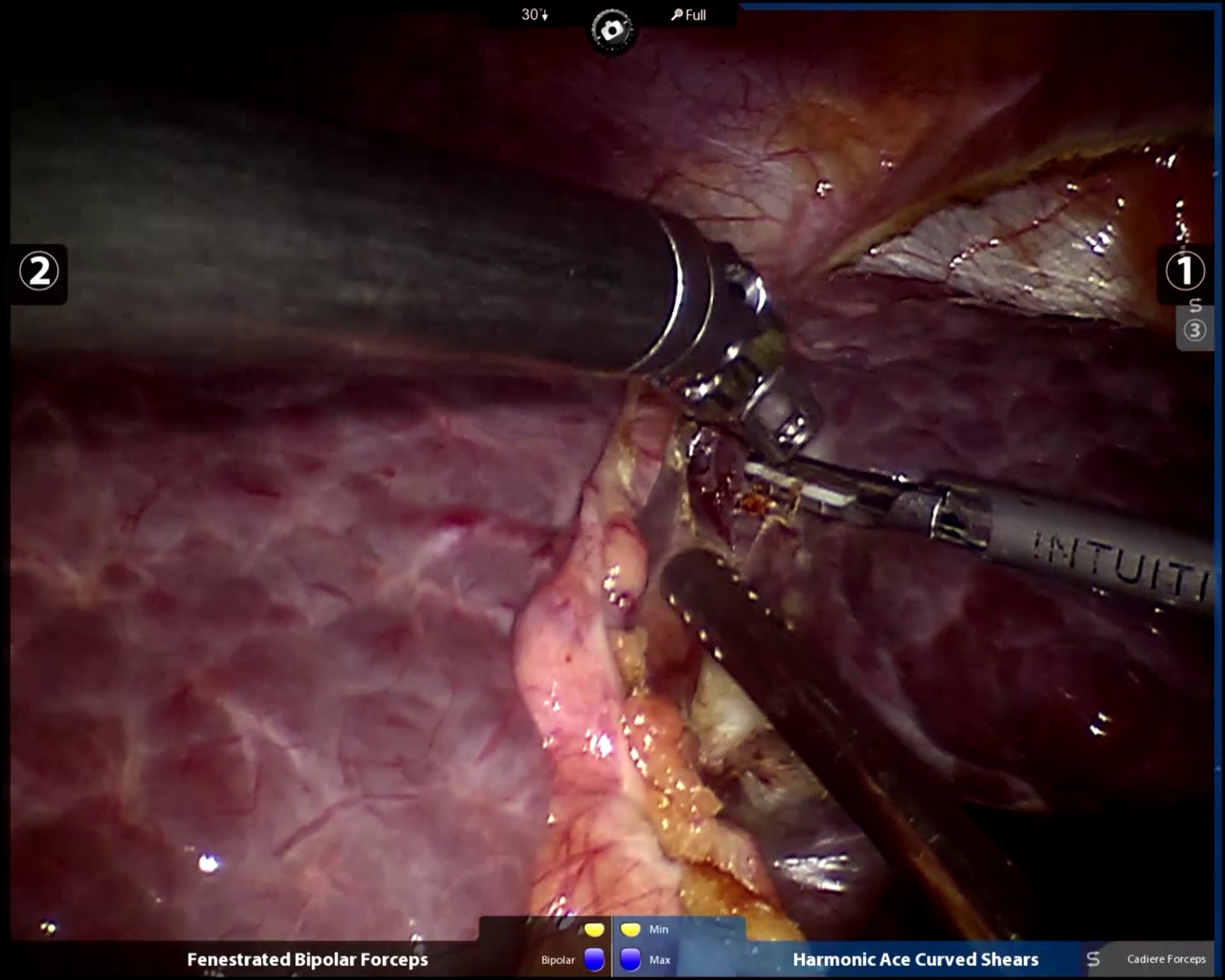}
\includegraphics[width=0.226\linewidth]{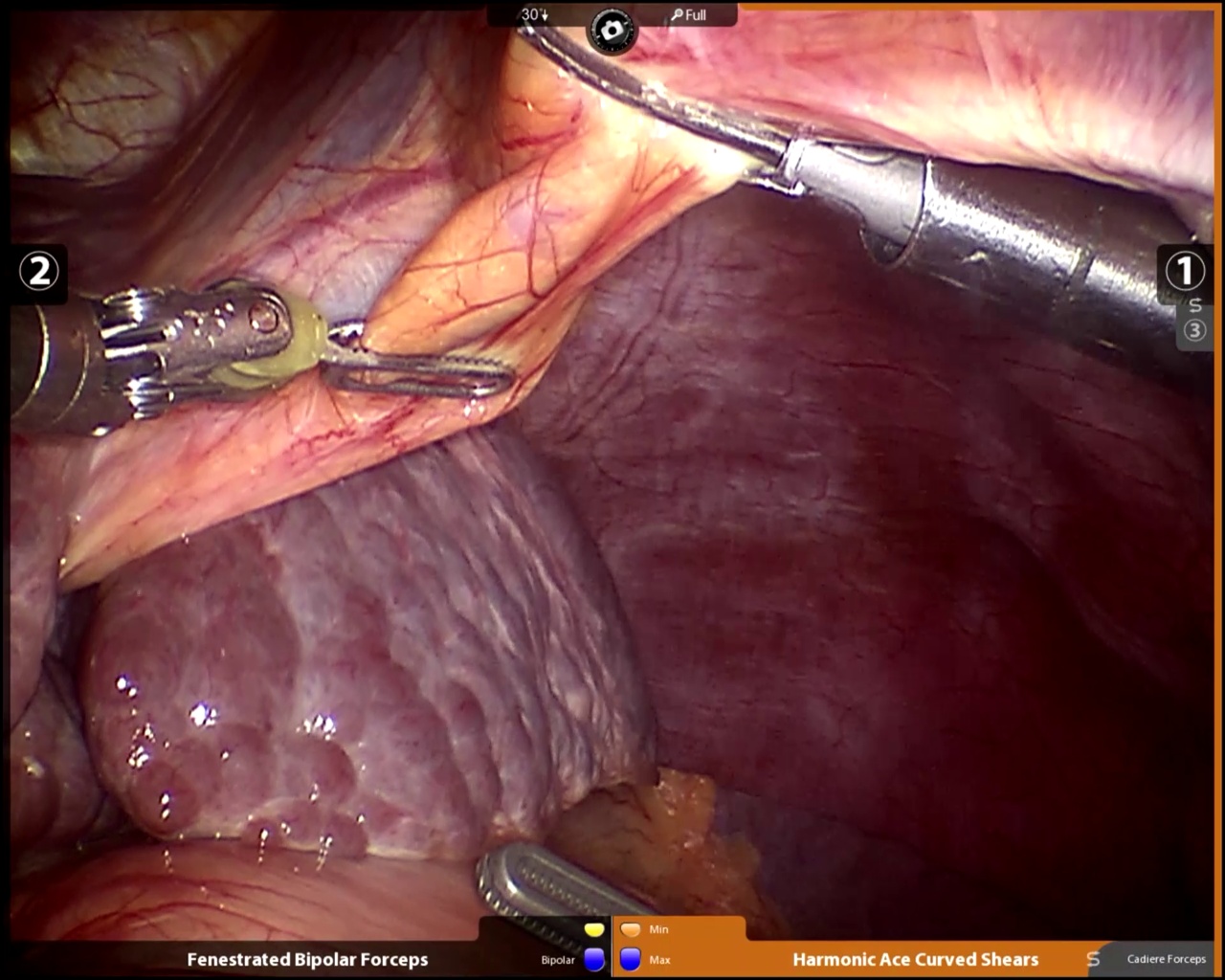}
\includegraphics[width=0.226\linewidth]{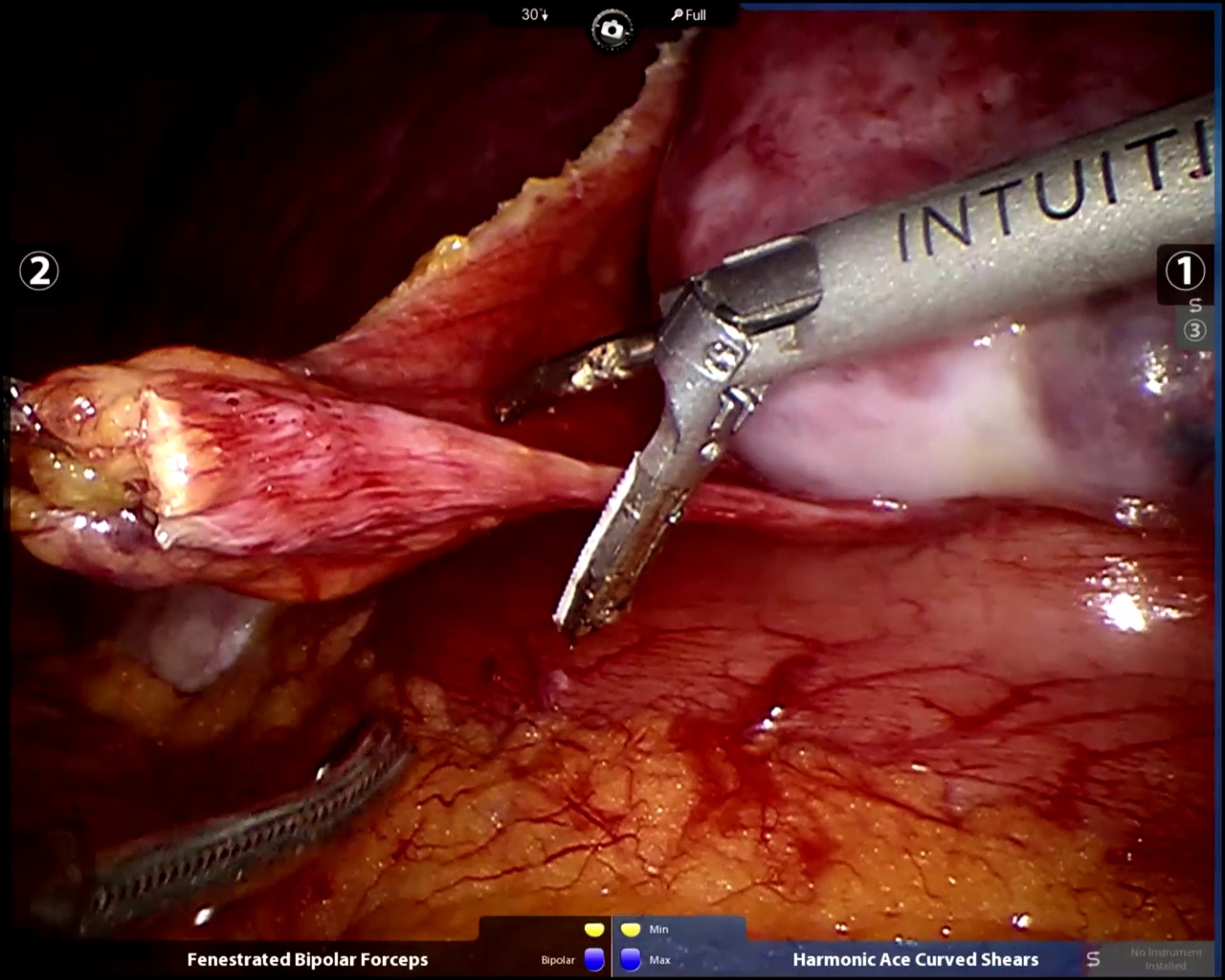}
\includegraphics[width=0.226\linewidth]{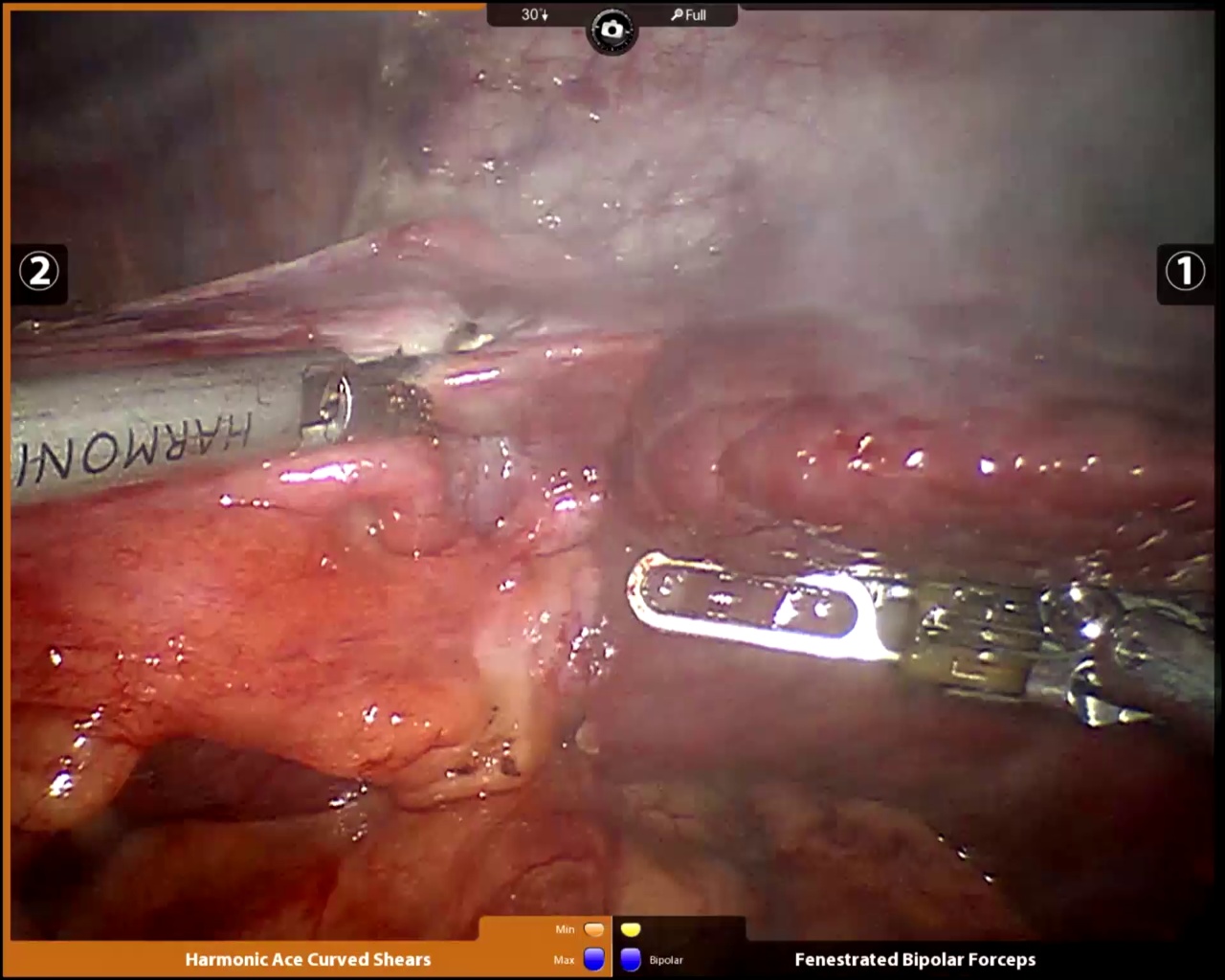}
}
\centerline{
\includegraphics[width=0.226\linewidth]{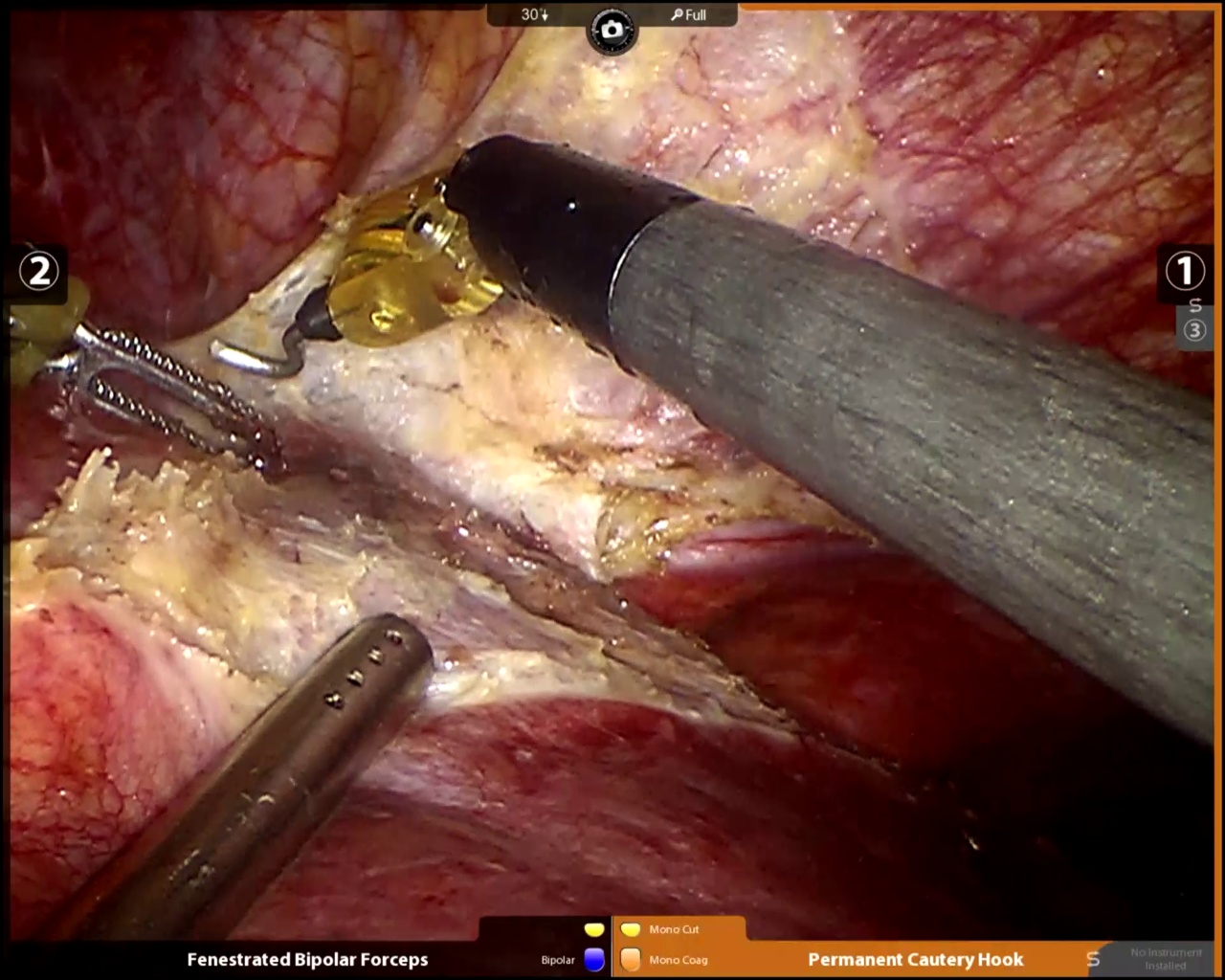}
\includegraphics[width=0.226\linewidth]{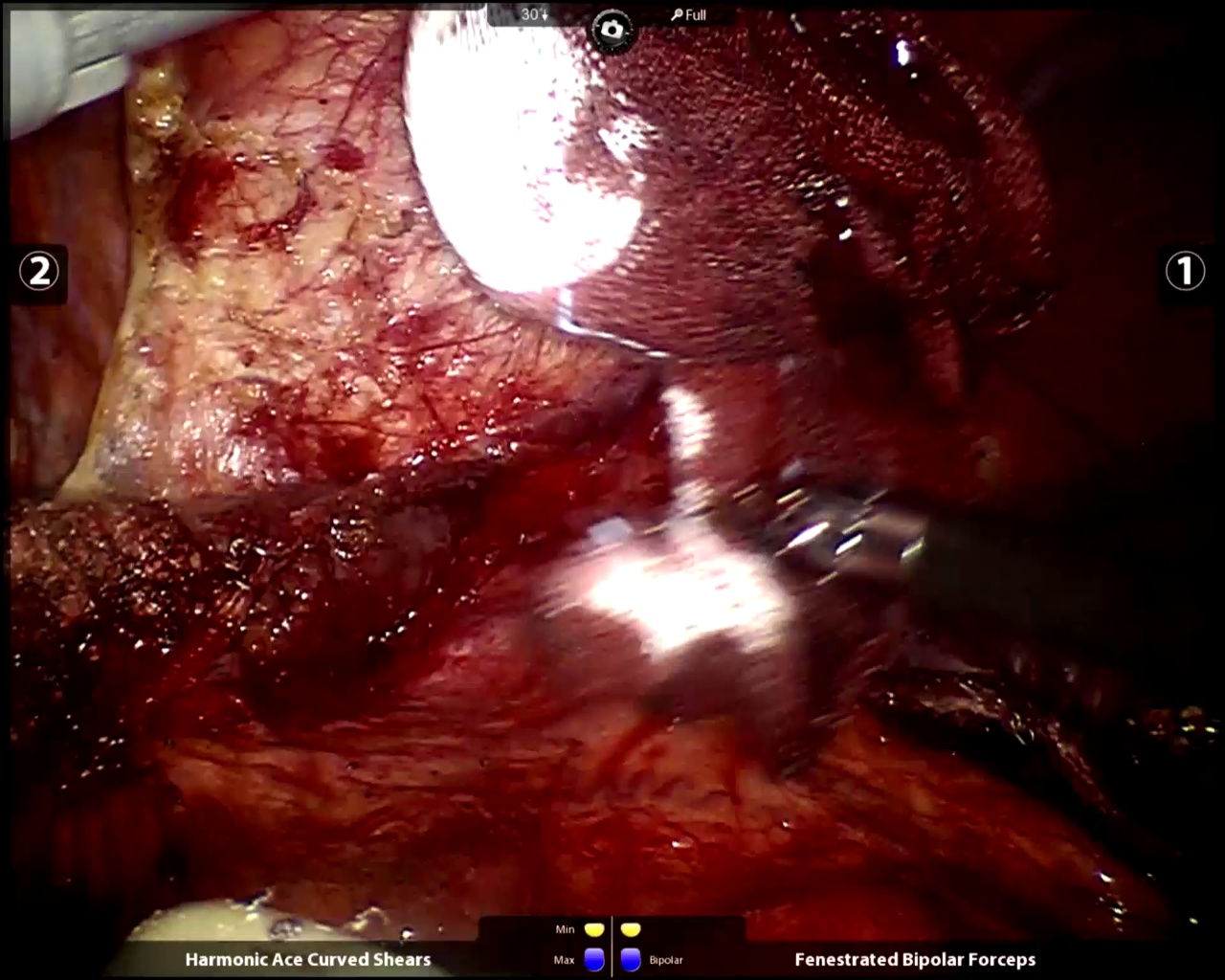}
\includegraphics[width=0.226\linewidth]{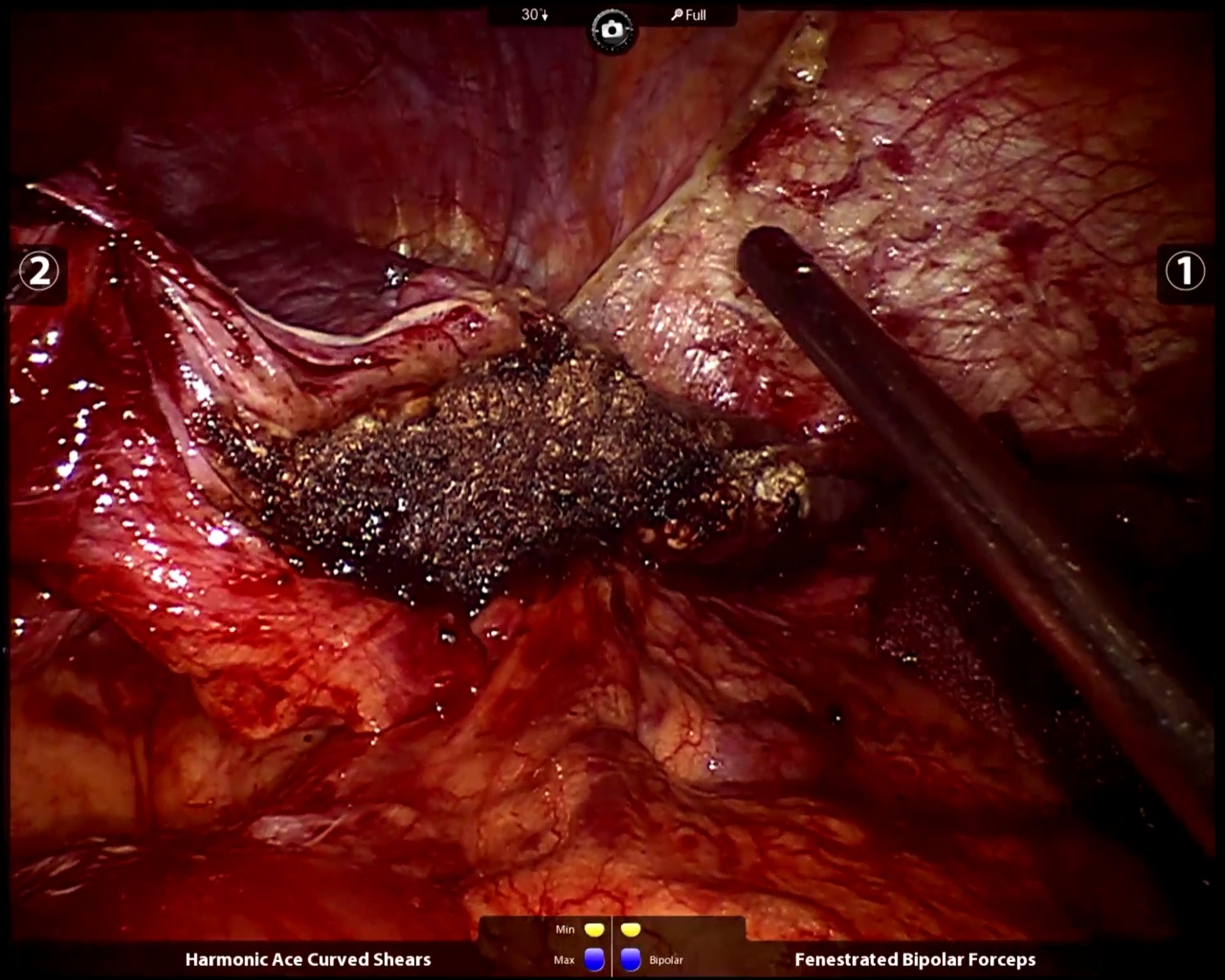}
\includegraphics[width=0.226\linewidth]{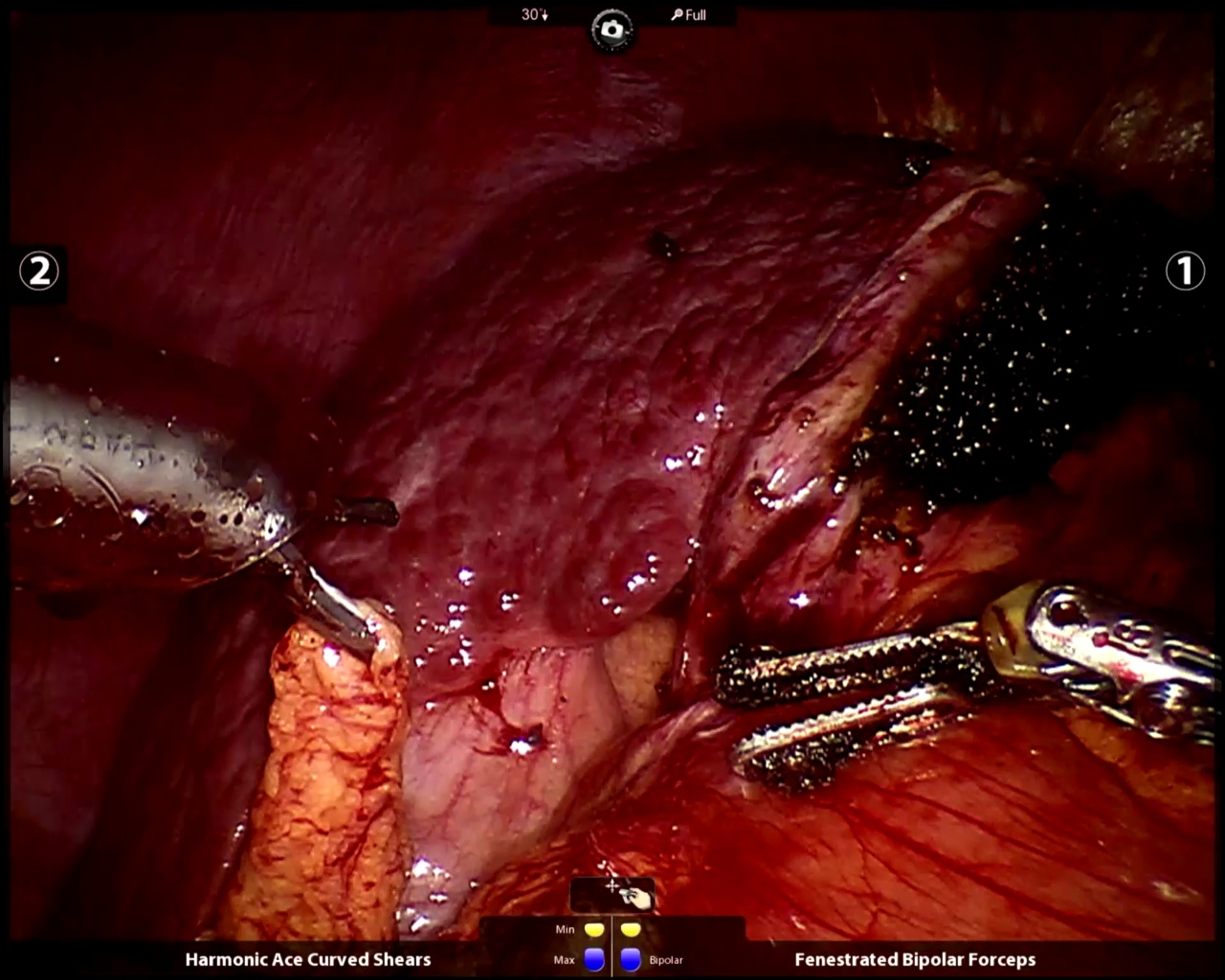}
}
\caption{Examples frames in the RLLS12M dataset. We collect the entire RLLS procedures in our dataset, involving all possible scenarios that happen in the clinical environment. The data frames not only contain regular endoscopic views, but also contain challenging scenarios for computer vision algorithms, such as smoke, specularity, texture-less view, and blood leaking. Note that the bottom right image is an example of an under-effective frame, where a surgeon manipulates the instruments without performing an effective action.}
\label{fig:dataset_samples}
\end{figure}

\begin{figure*}[t!]
\centerline{\includegraphics[width=\linewidth]{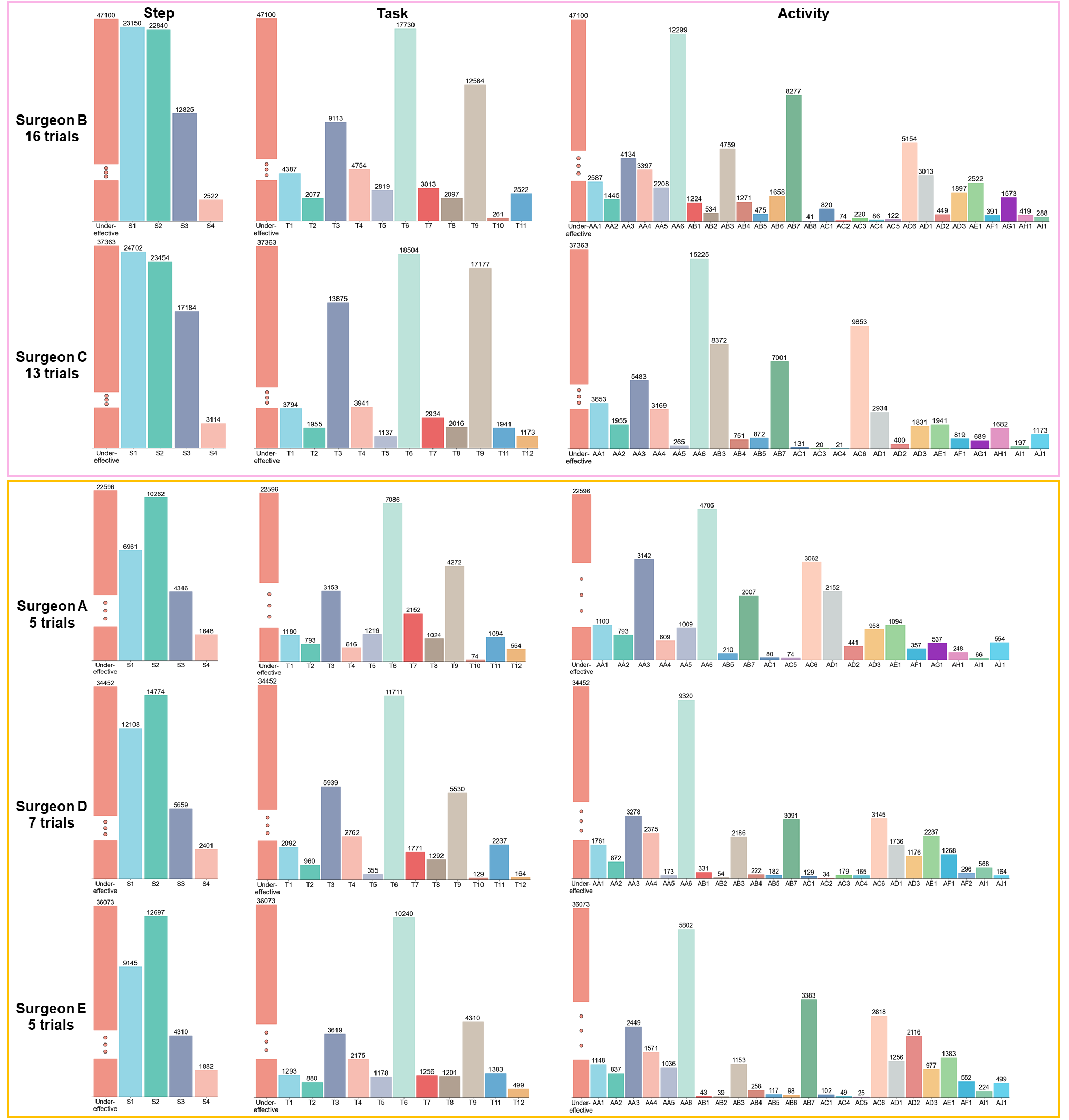}}
\caption{Hierarchical label distributions of surgeons in the \textit{RLLS} configuration. Training sample distributions of the two surgeons are in the pink rectangle. Testing ones are in the yellow rectangle. For each annotation type, we assign each label a specific color to better visualize the style preferences of experts.}
\label{fig:rlls_split}
\end{figure*}

\begin{figure*}[t!]
\centerline{\includegraphics[width=\linewidth]{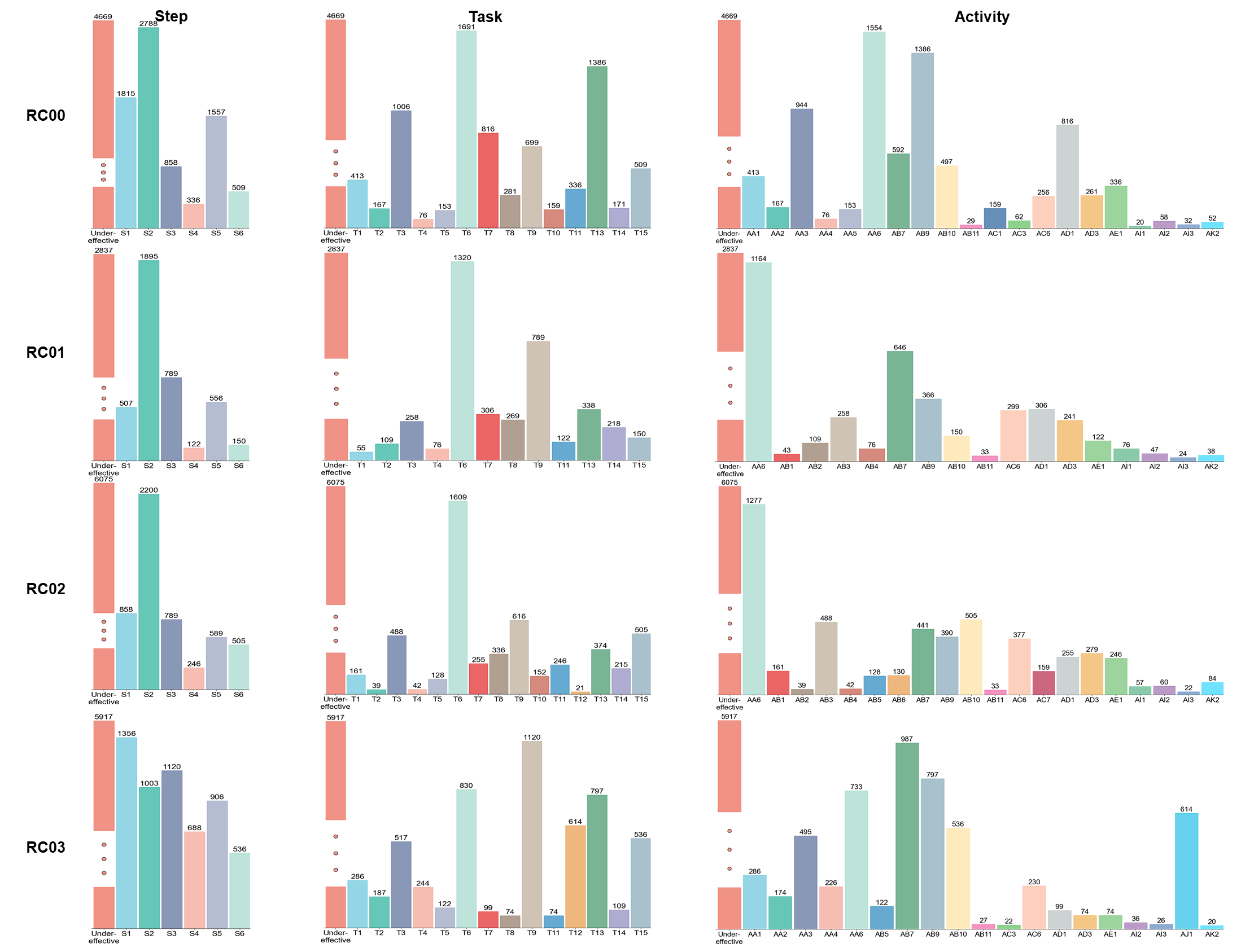}}
\caption{Label distribution of \textit{RLLS-w-RC} configuration.}
\label{fig:RLLS-w-RC_split}
\end{figure*}

\subsection{Loss Function}
Since the proposed RLLS12M dataset contains hierarchical annotations with one-to-many containment relationships and triplet activities, we follow ~\cite{nwoye2022rendezvous} to employ each supervised module with a cross-entropy loss so that the total loss for training is in a weighted sum representation:
\begin{equation}
\begin{aligned}
    L_{total} = \alpha_{S}L_{S} + \alpha_{T}L_{T} + \alpha_{\langle IAO\rangle}L_{\langle IAO\rangle} \\ + \alpha_{I}L_{I} + \alpha_{A}L_{A} + \alpha_{O}L_{O},
\end{aligned}
\label{eq:totalloss}
\end{equation}
where $L$ and $\alpha$ represent the loss function and its corresponding weight. All $\alpha$ values are simply set to 1.0 in all experiments to treat all tasks equally.

\section{Experiments}
\subsection{Data Preparation}
We decrease the frame resolution in the RLLS12M dataset to a 320 $\times$ 256 resolution for training efficiency. All images are selected from the image set by down-sampling them with a 5 fps sampling rate. 
We offer two experimental configurations based on the type of surgical workflow in the dataset: \textit{RLLS} and \textit{RLLS-w-RC}. The \textit{RLLS} configuration only has RLLS sequences, as shown in Fig.~\ref{fig:rlls_split}, including 29 videos from 2 expert surgeons. The test set consists of in total 17 RLLS videos from the other 3 expert surgeons; more details are in Table \ref{tab:data_split_rlls}. We name RLLS with robotic cholecystectomy \textit{RLLS-w-RC}, as shown in Fig.~\ref{fig:RLLS-w-RC_split}, which contains 4 videos of RLLS with RC from different experts to set up the experiment with leave-one-out-cross-validation. The statistical information of \textit{RLLS-w-RC} is in Table \ref{tab:data_split_RLLS-w-RC}. Note that each specialist has an operating style preference, which does not prevent them from strictly following standard procedures. This surgeon-level split makes our evaluation to focus on not only the prediction quality but also the understanding of varying surgical styles. The 4 RLLS with cholecystectomy videos have a partially overlapped domain knowledge to RLLS, aiming to evaluate the robustness of the proposed relation-based framework. 

\begin{table}[t]
    \scriptsize
	\setlength\tabcolsep{1.5pt}
	\begin{center}
		\caption{The statistics of data split for \textit{RLLS} configuration. The training set includes 29 videos from 2 surgeons and the testing set contains 21 videos from the other 3 surgeons. }
		\label{tab:data_split_rlls}
		\begin{tabular}{  l | l | r | r| r | r }
			\hline
			RLLS
			& Surgeons & Videos & Frames & Effective frames & Under-effective frames  \\ \hline
			\textbf{Training} & B,C & 29 & 267,631 & 153,097 & 114,534\\  
			\textbf{Testing} & A,D,E & 17 & 179,314 & \enspace86,193 & \enspace93,121\\
			\hline
			\textbf{Total} & A,B,C,D,E & 46 & 446,945 & 239,290 & 207,655\\

			    \hline
		\end{tabular}
	\end{center}
\end{table}

\begin{table}[t!]
    \scriptsize
	\setlength\tabcolsep{3.5pt}
	\begin{center}
		\caption{Data statistics of the \textit{RLLS-w-RC} configuration for leave-one-out cross-validation.}
		\label{tab:data_split_RLLS-w-RC}
		\begin{tabular}{  l | l | r | r | r  }
			\hline
			\textit{RLLS-w-RC}
			& Surgeons & Frames & Effective frames & Under-effective frames  \\ \hline
			\textbf{RC00} & C & 12,532 & \enspace7,863 & \enspace4,669\\
			\textbf{RC01} & B & \enspace6,856 & \enspace4,019 & \enspace2,837\\
			\textbf{RC02} & B & 11,262 & \enspace5,187 & \enspace6,075\\
			\textbf{RC03} & D & 11,526 & \enspace5,622 & \enspace5,917\\
			\hline
			\textbf{Total} & B,C,D  & 42,176 & 22,702 & 19,474\\

			    \hline
		\end{tabular}
	\end{center}
\end{table}

\begin{table*}[ht]
    \scriptsize
	\setlength\tabcolsep{4.8pt}
	\begin{center}
		\caption{The results of average precision on the \textit{RLLS} testing set. Results show that our method not only outperforms all baseline methods on primary tasks but also achieves the SOTA performance among existing representative methods. The symbol `-' denotes the unsupported performance measurements for a specific model. The comparison reference is MURPHY with CNN-LSTM backbone. Within each column, the {\bf best} and \underline{second-best} are highlighted. The \textcolor{blue}{blue} and \textcolor{red}{red} numbers indicate better and worse performance than the reference, respectively. }
		\label{tab:sap_results}
		\begin{tabular}{  l | ll | lll | lll   }
			\hline
			\multirow{2}{*}{{\textit{RLLS}}}
			&\multicolumn{2}{c}{{}}   
			& \multicolumn{3}{c}{{Average precision}}  
			&\multicolumn{3}{c}{{}}      \\ 
			\cline{2-9}
			& \enspace\enspace\enspace\enspace\textbf{SAP6} & \enspace\enspace\enspace\enspace\textbf{SAP3} & \enspace\enspace\enspace\enspace S & \enspace\enspace\enspace\enspace T & \enspace\enspace\enspace\enspace$\langle$IAO$\rangle$ & \enspace\enspace\enspace\enspace I & \enspace\enspace\enspace\enspace A & \enspace\enspace\enspace\enspace O \\ \hline
			CNN & 67.25$^{\pm2.61}$ \textcolor{red}{(\enspace1.28)} & 70.73$^{\pm2.07}$ \textcolor{red}{(\enspace2.43)} & 87.01 \textcolor{red}{(\enspace1.00)} & 69.03 \textcolor{red}{(\enspace1.15)} & 56.14 \textcolor{red}{(\enspace5.16)} & \underline{64.34} \textcolor{blue}{(\enspace0.12)}  & \underline{61.68} \textcolor{blue}{(\enspace0.53)} & 65.29 \textcolor{red}{(\enspace1.04)}\\
			
			
			CNN-LSTM & 66.30$^{\pm2.44}$ \textcolor{red}{(\enspace2.23)} & 70.38$^{\pm1.89}$ \textcolor{red}{(\enspace2.78)} & 85.36 \textcolor{red}{(\enspace2.65)} & 67.86 \textcolor{red}{(\enspace2.32)} & 57.93 \textcolor{red}{(\enspace3.37)} & 62.25 \textcolor{red}{(\enspace1.97)} & 60.43 \textcolor{red}{(\enspace0.72)} & 64.00 \textcolor{red}{(\enspace2.33)}\\
			
			ViT & 40.00$^{\pm5.23}$ \textcolor{red}{(28.53)} & 42.68$^{\pm4.36}$ \textcolor{red}{(30.48)} & 61.03 \textcolor{red}{(26.98)} & 37.77 \textcolor{red}{(32.41)} & 29.25 \textcolor{red}{(32.05)} & 42.50 \textcolor{red}{(21.72)}  & 37.71 \textcolor{red}{(23.44)} & 31.77 \textcolor{red}{(34.56)}\\
			
			Swin-Transformer & 51.12$^{\pm5.15}$ \textcolor{red}{(17.41)} & 54.77$^{\pm4.18}$ \textcolor{red}{(18.39)} & 72.69 \textcolor{red}{(15.32)} & 52.35 \textcolor{red}{(17.83)} & 39.25 \textcolor{red}{(22.05)} & 47.97 \textcolor{red}{(16.25)} & 46.33 \textcolor{red}{(14.82)} & 48.10 \textcolor{red}{(18.23)}\\
				\hline
			AttentionTripnet~\cite{nwoye2020recognition} & \enspace\enspace\enspace\enspace- & \enspace\enspace\enspace\enspace- & \enspace\enspace\enspace\enspace- & \enspace\enspace\enspace\enspace- & 36.42 \textcolor{red}{(24.88)}  & 50.01 \textcolor{red}{(14.21)} & 49.28 \textcolor{red}{(11.87)} & 51.92 \textcolor{red}{(14.41)}\\
			
			Rendezvous~\cite{nwoye2022rendezvous} & \enspace\enspace\enspace\enspace- & \enspace\enspace\enspace\enspace- & \enspace\enspace\enspace\enspace- & \enspace\enspace\enspace\enspace- & 49.20 \textcolor{red}{(12.10)} & 56.92 \textcolor{red}{(\enspace7.30)} & 55.39 \textcolor{red}{(\enspace5.75)} & 57.05 \textcolor{red}{(\enspace9.28)}\\
			
		    MTMS-TCN~\cite{ramesh2021multi} & \enspace\enspace\enspace\enspace- & \enspace\enspace\enspace\enspace- & 83.57 \textcolor{red}{(\enspace4.44)} & 63.60 \textcolor{red}{(\enspace6.58)} & \enspace\enspace\enspace\enspace- & \enspace\enspace\enspace\enspace- & \enspace\enspace\enspace\enspace- & \enspace\enspace\enspace\enspace-\\
		    
		    MTMS-TCN~\cite{ramesh2021multi} (full) & 65.35$^{\pm2.91}$ \textcolor{red}{(\enspace3.18)} & 68.81$^{\pm2.15}$ \textcolor{red}{(\enspace4.35)} & 85.18 \textcolor{red}{(\enspace2.83)} & 66.16 \textcolor{red}{(\enspace4.02)} & 55.11 \textcolor{red}{(\enspace6.19)} & 62.33 \textcolor{red}{(\enspace1.89)} & 60.16 \textcolor{red}{(\enspace0.99)} & 63.19 \textcolor{red}{(\enspace3.14)}\\
			    \hline
			MURPHY(CNN) & \textbf{69.24}$^{\pm1.96}$ \textcolor{blue}{(\enspace0.71)} & \underline{73.10}$^{\pm1.68}$ \textcolor{red}{(\enspace0.06)} & \underline{87.76} \textcolor{red}{(\enspace0.25)} & \textbf{70.57} \textcolor{blue}{(\enspace0.39)} & \underline{60.96} \textcolor{red}{(\enspace0.34)} & \textbf{65.58} \textcolor{blue}{(\enspace1.36)} & \textbf{63.40} \textcolor{blue}{(\enspace2.25)} & \textbf{67.18} \textcolor{blue}{(\enspace0.85)}\\
			MURPHY(CNN-LSTM) & \underline{68.53}$^{\pm2.74}$ & \textbf{73.16}$^{\pm2.27}$ & \textbf{88.01} & \underline{70.18} & \textbf{61.30} & 64.22 & 61.15 & \underline{66.33}\\
			\hline
		\end{tabular}
	\end{center}
\end{table*}

\begin{table*}[t!]
    \scriptsize
	\setlength\tabcolsep{3.0pt}
	\begin{center}
		\caption{The results of mean average precision on each surgeon's testing set in the \textit{RLLS} experiment. Results show that our method not only outperforms baseline methods but also achieves SOTA performance among existing representative methods. The larger value indicates better performance. Within each column, the {\bf best} and \underline{second-best} are highlighted.}
		\label{tab:surgeon_results}
		\begin{tabular}{  l | rrrrrr | rrrrrr | rrrrrr }
			\hline
			\multirow{2}{*}{{\textit{RLLS}}}
			&\multicolumn{6}{c|}{{Surgeon A}}   
			& \multicolumn{6}{c|}{{Surgeon D}}  &\multicolumn{6}{c}{{Surgeon E}}      \\ 
			\cline{2-19}
			& S & T & $\langle$IAO$\rangle$ & I & A & O 
			& S & T & $\langle$IAO$\rangle$ & I & A & O 
			& S & T & $\langle$IAO$\rangle$ & I & A & O \\ 
			    \hline
			CNN & 87.82 & 73.27 &  59.28  & \underline{68.11} & \underline{65.15} & 72.00 & 87.42 & 64.42 & 53.50 & \underline{62.66} & 60.41 & 63.52 & 85.78 & 69.40 & {55.64} & \textbf{66.25} & \underline{59.48} & \underline{60.36}\\
			CNN-LSTM & 86.54 & 71.98 & 60.19 & 64.48 & 63.42 & 69.76 & 86.91 & 65.30 & 57.37 & 62.01 & \underline{60.45} & 64.41 & 82.62 & 66.31 & 56.22 & 60.25 & 57.41 & 57.84 \\
			ViT & 64.97 & 44.84 & 36.64 & 46.35 & 46.94 & 42.26 & 60.29 & 33.00 & 27.30 & 42.57 & 37.25 & 31.08 & 57.82 & 35.46 & 23.82 & 38.57 & 28.94 & 21.96 \\
			Swin-Transformer & 77.80 & 60.65 & 43.35 & 52.70 & 55.42 & 59.22 & 71.23 & 46.78 & 40.02 & 47.62 & 43.97 & 44.83 & 69.05 & 49.62 & 34.39 & 43.59 & 39.60 & 40.25 \\
			    \hline
		   
			AttentionTripnet~\cite{nwoye2020recognition} & - & - & 38.03 & 45.71 & 49.13 & 52.51 & - & - & 36.63 & 52.99 & 50.81 & 55.12 & - & - & 34.60 & 51.32 & 47.90 & 48.12  \\
			Rendezvous~\cite{nwoye2022rendezvous} & - & - & 50.31 & 61.12 & 59.97 & 61.40 & - & - & 49.13 & 55.53 & 52.71 & 57.20 & - & - & 48.16 & 54.12 & 53.50 & 52.56 \\
			MTMS-TCN~\cite{ramesh2021multi} & 84.07 & 67.81 & - & - & - & - & 85.16 & 60.77 & - & - & - & - & 81.47 & 62.23 & - & - & - & - \\
		    MTMS-TCN~\cite{ramesh2021multi} (full) & 85.93 & 70.09 & 59.45 & 65.97 & 64.20 & 70.40 & 86.16 & 63.21 & 53.72 & 61.19 & 59.17 & 62.16 & 83.44 & 65.17 & 52.16 & 59.83 & 57.12 & 57.01 \\
		    	\hline
			MURPHY(CNN) & \underline{88.19} & \textbf{73.91} & \underline{64.05} & 66.80 & \textbf{65.21} & \textbf{73.80} & \underline{88.24} & \underline{66.81} & \underline{59.16} & \textbf{64.56} & \textbf{62.25} & \underline{64.71} & \underline{86.86} & \textbf{71.00} & \textbf{59.68} & \underline{65.37} & \textbf{62.73} & \textbf{63.02}\\
			MURPHY(CNN-LSTM) & \textbf{88.61} & \underline{73.72} & \textbf{66.80} & \textbf{68.45} & {64.10} & \underline{72.43} & \textbf{88.27} & \textbf{67.33} & \textbf{59.19} & 62.28 & 59.98 & \textbf{66.25} & \textbf{87.15} & \underline{69.48} & \underline{57.91} & 61.93 & 59.36 & 60.31\\
			\hline
		\end{tabular}
	\end{center}
\end{table*}

\subsection{Implementation Details}
The proposed model is implemented with the PyTorch framework. 
We apply a Stochastic Gradient Descent optimization with default parameters to train the proposed model. The learning rate is initialized to 0.01 with a decay rate of 0.99 for all methods. Our model is trained with around 50 epochs on an NVIDIA A100 GPU in the \textit{RLLS} setup. Full training takes approximately 75 hours. In the \textit{RLLS-w-RC} setup, we employ the pre-trained model to fine-tune 10 epochs for evaluation. The total computation resource for the model, input data, temporal cache, and other auxiliary buffers of batch size 16 is under 10 GB. The total parameters of our configuration (with a 2-layered R-GCN module~\cite{li2018deeper}) are about 31.6M.

\subsection{Evaluation Metrics}
\subsubsection{Mean Average Precision}
The mean average precision (mAP) is the metric for performance evaluation, indicating the reliable level of model predictions. We define an averaged mAP for step, task, and triplet activity as summarized average precision for 3 primary tasks (SAP3) metrics to better describe the overall performance of the surgical analysis model. \textbf{SAP3 is the most important metric}, which reports the overall performance of all primary tasks (step, task, activity) in our surgical interpretation. Similarly, we define another metric, SAP6, to evaluate the average precision for 3 primary tasks and 3 auxiliary tasks (instrument, action, object). 

\subsubsection{Edit Distance}
We use edit distance as the quantitative evaluation metric for prediction orders~\cite{stauder2016tum}. This metric describes the distance between ground truth and predicted label sequences without considering the timing. Since the minimum temporal annotation and the minimum under-effective time interval are 2 and 5 seconds, we select 10 and 25 frames tolerance to evaluate the performance for reasonable comparisons.

\subsection{Comparison Methods}
\subsubsection{Representative Baselines}
We select the popular architectures that are aforementioned in related works for comparisons, including ResNet50~\cite{he2016deep}, CNN-LSTM~\cite{yengera2018less}, Vision Transformer (ViT)~\cite{dosovitskiy2020vit}, and Swin-Transformer~\cite{liu2021Swin}. All these networks can be used as the baseline in our proposed model. We only report and discuss the MURPHY with CNN-LSTM and CNN backbones based on the observations from recent work~\cite{jin2021temporal, ramesh2021multi}.

\subsubsection{Existing Methods}
We also compare some representative works in surgical procedure analysis, including state-of-the-art (SOTA) algorithms for step or task recognition and triplet activity recognition. More specifically, we compare our method with the Rendezvous~\cite{nwoye2022rendezvous} and AttentionTripnet~\cite{nwoye2020recognition} methods for triplet activity recognition, and MTMS-TCN~\cite{ramesh2021multi} for step and task recognition. These comparisons can provide some observations on the better capability of the proposed model than existing works.

\section{Results}

\subsection{Results of \textit{RLLS} Configuration}
The results of baseline comparisons are shown in Table ~\ref{tab:sap_results}. Our MURPHY outperforms all comparison methods, achieving superior results on all primary tasks (step, task, activity). Results also show that our method achieve compatible results on most of the auxiliary tasks (instrument, action, object). Our method achieves the best and the second-best results on both SAP3 and SAP6. Specifically, MURPHY outperforms CNN and CNN-LSTM baselines, which are the top 2 baselines on the SAP3 metric. These observations show that the proposed relation-based modules facilitate workflow prediction on all primary tasks. Similar trends are observed in the results of the SAP6 metric as well. Note that both CNN and CNN-LSTM-based MURPHY models achieve superior mAP results than other methods, indicating that our relation-based modules facilitate the representation power for classification with the robustness of backbone selection.

\begin{table*}[ht]
    \scriptsize
	\setlength\tabcolsep{10.2pt}
	\begin{center}
		\caption{The results of edit distance. Under-effective frames not only contain the absence of instruments but also surgical instrument presence with ineffective behaviors, resulting in the increasing difficulty of distinguishing abundant annotation categories for neural networks. Within each type of RLLS annotation, the {\bf best} and \underline{second-best} are highlighted.
        }
		\label{tab:ed_results}
		\begin{tabular}{  l | l | rr | rr | rr | rr  }
			\hline
			\multirow{2}{*}{{\textit{RLLS}}}
			&\multicolumn{1}{c|}{\multirow{2}{*}{{Method}}}  
			&\multicolumn{2}{c|}{{Average}}   
			&\multicolumn{2}{c|}{{Surgeon A}}   
			&\multicolumn{2}{c|}{{Surgeon D}}  
			&\multicolumn{2}{c}{{Surgeon E}}      \\ 
			\cline{3-10}
			&  & ED@10 & ED@25  & ED@10 & ED@25  & ED@10 & ED@25  & ED@10 & ED@25 \\ \hline
			\multirow{6}{*}{{\enspace S}} 
			& CNN  & \enspace26.21 & \enspace38.19  & \enspace17.20 & \enspace27.60 & \enspace24.43 & \enspace35.57 & \enspace37.00 & \enspace\textbf{51.40}\\
			& CNN-LSTM & \enspace22.42  & \enspace\underline{37.20}  & \enspace14.20  & \enspace\textbf{23.00} & \enspace\underline{19.86} & \enspace\underline{34.00} & \enspace\textbf{33.20} & \enspace54.60\\
            & MTMS-TCN~\cite{ramesh2021multi} & \enspace27.83 & \enspace39.50 & \enspace17.80 & \enspace25.60 & \enspace24.29 & \enspace36.71 & \enspace41.40 & \enspace56.20\\
            & MTMS-TCN~\cite{ramesh2021multi} (full) & \enspace26.55  & \enspace39.31  & \enspace\underline{13.40}  & \enspace26.20  & \enspace24.86 & \enspace35.14 & \enspace41.40 & \enspace56.60\\           
           	& MURPHY(CNN) & \enspace\underline{21.89}  & \enspace39.01  & \enspace\textbf{10.80}  & \enspace25.80 & \enspace21.28 & \enspace38.42 & \enspace\underline{33.60} & \enspace52.80\\ 
                & MURPHY(CNN-LSTM) & \enspace\textbf{20.49}  & \enspace\textbf{34.95}  & \enspace\underline{13.40}  & \enspace\underline{24.20}   & \enspace\textbf{14.28} & \textbf{\enspace28.85} & \enspace33.80 & \enspace\underline{51.80}\\

            \hline
            
            \multirow{6}{*}{{\enspace T}} 
            & CNN & \enspace31.45 & \enspace40.48 & \enspace18.00 & \enspace26.00 & \enspace28.14 & \enspace36.43 & \enspace48.20 & \enspace\underline{59.00}\\
            & CNN-LSTM & \enspace\underline{29.14} & \enspace\underline{39.90} & \enspace18.40 & \enspace26.20 & \enspace\underline{24.43} & \enspace34.29 & \enspace\textbf{44.60} & \enspace59.20\\
            & MTMS-TCN~\cite{ramesh2021multi} & \enspace33.03 & \enspace41.97 & \enspace17.40 & \enspace26.22 & \enspace29.29 & \enspace38.71 & \enspace52.40 & \enspace61.00\\
            & MTMS-TCN~\cite{ramesh2021multi} (full) & \enspace31.23  & \enspace40.00  & \enspace17.00  & \enspace26.60   & \enspace27.29 & \enspace\underline{34.00} & \enspace49.40 & \enspace59.40\\
            & MURPHY(CNN) & \enspace29.89 & \enspace40.63  & \enspace\textbf{14.40}  & \enspace\textbf{23.80}   & \enspace28.28 & \enspace37.28 & \enspace\underline{47.00} & \enspace60.80\\
            & MURPHY(CNN-LSTM) & \enspace\textbf{28.86}  & \enspace\textbf{37.78}  & \enspace\underline{16.80}  & \enspace\underline{24.80}   & \enspace\textbf{22.57} & \textbf{\enspace31.14} & \enspace47.20 & \enspace\textbf{57.40}\\
            
            \hline
            
            \multirow{7}{*}{{$\langle$IAO$\rangle$}} 
            & CNN & \enspace31.65 & \enspace52.95 & \enspace19.40 & \enspace33.20 & \enspace30.14 & \enspace46.86 & \enspace\underline{45.40} & \enspace78.80\\
            & CNN-LSTM & \enspace\underline{31.48} & \enspace51.28 & \enspace\underline{19.80} & \enspace32.80  & \enspace27.43 & \enspace47.43 & \enspace47.20 & \enspace73.60 \\
            & AttentionTripnet~\cite{nwoye2020recognition} & \enspace35.97 & \enspace58.98  & \enspace23.80 & \enspace39.60 & \enspace31.71 & \enspace52.14 & \enspace52.40 & \enspace85.20\\
            & Rendezvous~\cite{nwoye2022rendezvous} & \enspace32.92 & \enspace53.54 & \enspace21.40 & \enspace36.00 & \enspace\underline{26.57} & \enspace47.42 & \enspace50.80 & \enspace77.20\\
            & MTMS-TCN~\cite{ramesh2021multi} (full) & \enspace32.89  & \enspace51.62  & \enspace\textbf{18.80}  & \enspace32.20 & \enspace29.86 & \enspace44.86 & \enspace50.00 & \enspace77.80\\
            & MURPHY(CNN) & \enspace34.24 & \enspace\underline{47.35} & \enspace25.00 & \enspace\underline{27.80} & \enspace30.71 & \enspace\underline{43.85} & \enspace47.00 & \enspace\underline{70.40}\\
            & MURPHY(CNN-LSTM) & \enspace\textbf{30.04} & \enspace\textbf{46.12} & \enspace\underline{19.80} & \enspace\textbf{26.80} & \enspace\textbf{25.71} & \enspace\textbf{41.57} & \enspace\textbf{44.60} & \enspace\textbf{70.00}\\

            \hline
		\end{tabular}
	\end{center}
\end{table*}

\begin{table*}[t]
    \scriptsize
	\setlength\tabcolsep{7.0pt}
	\begin{center}
	\caption{The results of average precision for the \textit{RLLS-w-RC} test set. The symbol `-' denotes the unsupported performance measurements for a specific model. Results show that our method attains much better results than other methods, realizing a faster learning ability to transfer the knowledge to new trials. }
	\label{tab:lc_cv_results}
	\begin{tabular}{  l | ll | lll | lll  }
		\hline
		\multirow{2}{*}{{\textit{RLLS-w-RC}}}
		&\multicolumn{2}{c}{{}}   
		& \multicolumn{3}{c}{{Average precision}}  
		&\multicolumn{3}{c}{{}}      \\ 
		\cline{2-9}
		& \textbf{SAP6} & \textbf{SAP3} & S & T & $\langle$IAO$\rangle$ & I & A & O \\ \hline
		CNN & 59.17$^{\pm6.17}$ & 60.81$^{\pm5.58}$ & 74.67$^{\pm5.37}$ & 57.90$^{\pm6.12}$ & 49.86$^{\pm5.68}$ & 62.21$^{\pm6.95}$  & 58.90$^{\pm7.41}$ & 51.46$^{\pm6.89}$\\
		CNN-LSTM & 60.19$^{\pm7.23}$ & 61.73$^{\pm7.31}$ & 73.98$^{\pm8.44}$ & \underline{60.27}$^{\pm6.93}$ & 50.94$^{\pm6.78}$ & \underline{63.40}$^{\pm7.27}$ & \underline{59.32}$^{\pm8.31}$ & 52.63$^{\pm6.99}$\\
		ViT & 39.17$^{\pm2.71}$ & 40.11$^{\pm3.54}$ & 56.23$^{\pm4.60}$ & 34.76$^{\pm2.75}$ & 29.36$^{\pm3.54}$ & 45.43$^{\pm2.72}$ & 39.98$^{\pm2.68}$ & 29.24$^{\pm1.80}$\\
		Swin-Transformer & 47.83$^{\pm5.78}$ & 48.74$^{\pm5.11}$ & 64.98$^{\pm6.88}$ & 44.59$^{\pm5.34}$ & 36.66$^{\pm5.13}$ & 53.83$^{\pm7.20}$ & 48.90$^{\pm7.73}$ & 38.05$^{\pm5.31}$\\
			\hline
		AttentionTripnet~\cite{nwoye2020recognition} & - & - & - & - & 28.93$^{\pm4.08}$  & 46.76$^{\pm6.28}$ & 46.42$^{\pm6.25}$ & 35.94$^{\pm3.57}$\\
		Rendezvous~\cite{nwoye2022rendezvous} & - & - & - & - & 43.80$^{\pm3.43}$ & 58.36$^{\pm7.08}$ & 53.45$^{\pm7.74}$ & 45.02$^{\pm5.42}$\\
	    MTMS-TCN~\cite{ramesh2021multi} & - & - & 69.74$^{\pm5.60}$ & 50.66$^{\pm5.31}$ & - & - & - & -\\
	    MTMS-TCN~\cite{ramesh2021multi} (full) & 56.61$^{\pm5.02}$ & 58.67$^{\pm4.19}$ & 71.26$^{\pm3.49}$ & 56.04$^{\pm5.54}$  & 48.71$^{\pm3.96}$ & 59.14$^{\pm7.10}$ & 55.41$^{\pm7.82}$ & 49.10$^{\pm5.12}$\\
		    \hline

		MURPHY(CNN) & \textbf{61.45}$^{\pm5.22}$ & \textbf{64.08}$^{\pm5.50}$ & \textbf{77.45}$^{\pm6.02}$ & \textbf{62.26}$^{\pm5.23}$ & \textbf{52.53}$^{\pm5.41}$ & 63.22$^{\pm6.40}$ & \textbf{59.76}$^{\pm6.92}$ &\textbf{53.50}$^{\pm4.45}$\\
		MURPHY(CNN-LSTM) & \underline{61.25}$^{\pm5.45}$ & \underline{62.38}$^{\pm7.18}$ & \underline{75.33}$^{\pm6.84}$ & 59.39$^{\pm7.53}$ & \underline{52.44}$^{\pm7.39}$ & \textbf{68.98}$^{\pm6.25}$ & 58.74$^{\pm8.41}$ & \underline{52.64}$^{\pm8.15}$\\
		\hline
	\end{tabular}
	\end{center}
\end{table*}

We observe a similar trend in the comparison results of SOTA methods, as shown in Table \ref{tab:sap_results} and Table \ref{tab:surgeon_results}. AttentionTripnet and Rendezvous can hardly find compatible results with our method due to the lack of supervision from hierarchical annotations and the lack of utilizing relational information. Our method produces better prediction than the original MTMS-TCN because MURPHY systematically takes advantage of hierarchical annotations and relations. Because MTMS-TCN only concerns recognition of step and task, we extend it to recognize activity to provide a fair, full comparison with a representative multi-task method. The result shows that enriching the prediction tasks facilitate the performance by learning discriminative features from different supervision. Our method outperforms the MTMS-TCN with full tasks on the SAP3 metric and achieves superior prediction performance on all other tasks, suggesting that our relation-based module can effectively mine discriminative patterns in the relational information.

The results in Table \ref{tab:surgeon_results} report the detailed prediction performance of each surgeon in the \textit{RLLS} testing set. Our method achieves the SOTA performance on primary tasks in most cases which have a consistent trend with the previous summative results. The baseline methods prefer the easy tasks such as step recognition by sacrificing the performance on activity recognition while MTMS-TCN with full tasks achieves close performance but suffers performance degradation on analyzing the performance of Surgeon E. Moreover, our method consistently improves activity recognition in all three surgeons' procedures. These concrete observations further suggest that relations from hierarchical annotations provide useful information for surgical workflow analysis. 

To quantify the quality of the predicting order, we report the edit distance with relaxation for the testing set of \textit{RLLS} configuration in Table ~\ref{tab:ed_results} to evaluate the order prediction quality. Note that MURPHY achieves competitive edit distance results while preserving superior mAP results as reported in Table ~\ref{tab:sap_results}. This joint observation manifests that our method can recognize more challenging scenarios than other methods by considering the preservation of the sequential order using the relational feature. From the average results, CNN-LSTM-based MURPHY achieves the best edit distance in ED@10 and ED@25 on predicting all types of orders among these methods, improving the performance that inherits from the CNN-LSTM baseline. CNN-based MURPHY achieves compatible results in those order predictions. It improves activity order inference on ED@25 while maintaining similar performance to the CNN baseline on other tasks, indicating that the relational information makes positive efforts to order maintenance. Our MURPHY achieves competitive performance on the summary column for primary tasks, manifesting that our method can balance performance among hierarchical tasks without leaning too much on relatively easy classification tasks.

\subsection{Results of \textit{RLLS-w-RC} Configuration}
We observe that MURPHY outperforms both the baselines and the representative methods from the results of \textit{RLLS-w-RC} in Table \ref{tab:lc_cv_results}, which has the same trend as in the \textit{RLLS} setup. We notice that CNN and its variants still achieve better results than Transformers. MURPHY configurations outperform all other methods on most of the primary tasks and auxiliary tasks, especially for CNN-based MURPHY. Specifically, MURPHY achieves the top two performance on SAP6 and SAP3, indicating better knowledge transferring among comparison methods in the \textit{RLLS-w-RC} experiment. We observe that the varying preference of styles in Fig.~\ref{fig:RLLS-w-RC_split} that increases the difficulty of transfer learning, whereas MURPHY achieves better standard deviations than the corresponding baseline methods on SAP6 and SAP3, manifesting that MURPHY has a relatively stable adaptation to a similar surgical procedure with partially new annotations. These observations show that the effectiveness of the proposed relation modules, empowering MURPHY to achieve better transfer learning results.

\begin{figure*}[]
\centerline{
\includegraphics[width=\linewidth]{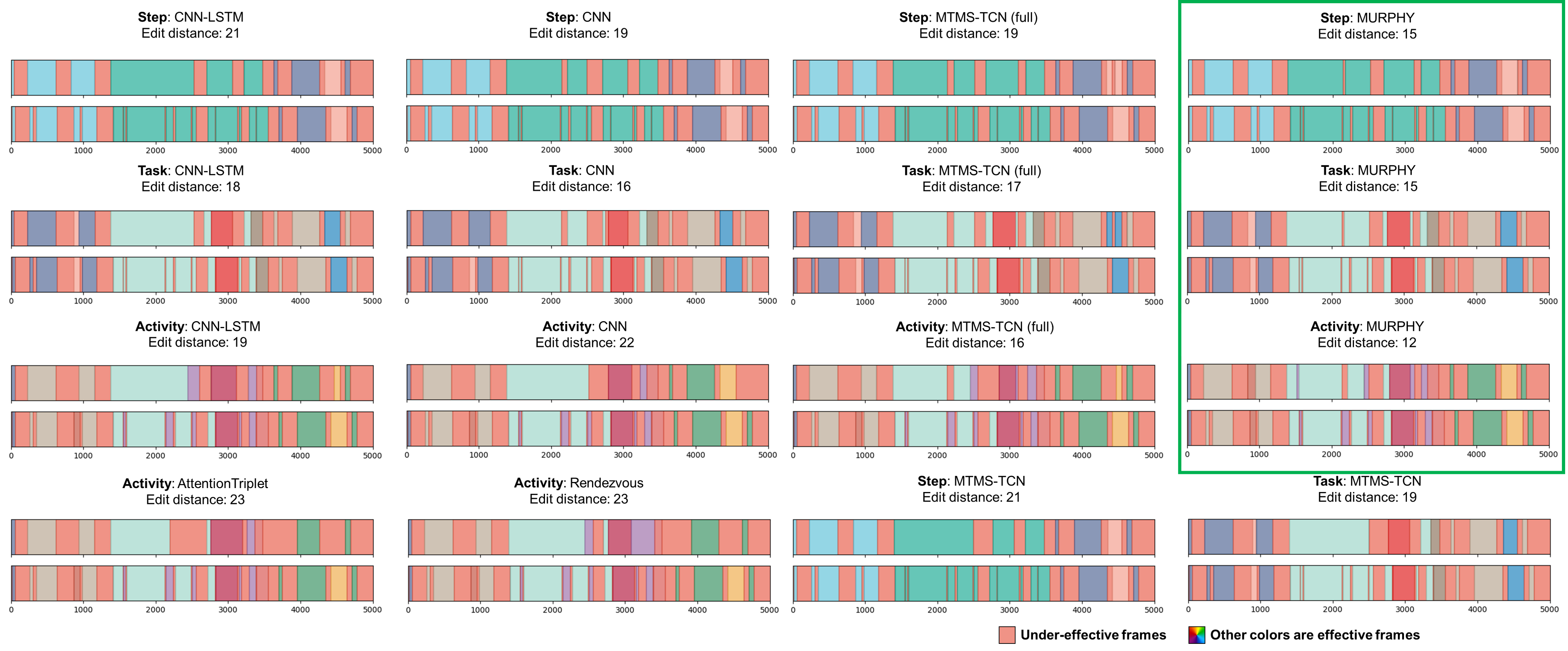}
}
\caption{Surgical workflow predictions on a 16-minute (5000 frames) long RLLS sequence with a minimal effective interval that is larger than 5 seconds, including representative comparison methods and MURPHY. The ground truth sequence is a challenging example containing abundant annotated temporal fragments in varying lengths. Results show that our prediction achieves the best results on edit distance (ED@25) than other methods. For each pair of colored sequences, the upper row is the inference result and the lower one is the ground truth. The smaller edit distance indicates better prediction on segment-level orders.}
\label{fig:qualitative_results2}
\end{figure*}

\subsection{Ablation Study}
To verify the effectiveness of the proposed modules, we conduct an ablation study that uses the CNN-LSTM backbone with multi-task classifiers as the reference method. This implementation based on MURPHY is different from the CNN-LSTM baseline method. The results in Table \ref{tab:ablation_results} show that both the R-GCN and HRCA modules are effective to primary task recognition. In addition, we observe that these relation learning modules make better improvement by jointly applying to the baseline model, suggesting that inter-relations and intra- relations can both provide useful information from different aspects. We find that MURPHY with the R-GCN module comprehensively improves the performance on all tasks because the R-GCN module learns the relationship from intra-relations and aggregates the relational features across inter-relations, encouraging the representation with more considerations on varying categorical labels and annotation types.

\begin{table}[t!]   
    \notsotiny
	\setlength\tabcolsep{3.6pt}
	\begin{center}
		\caption{The results of the ablation study on the RLLS testing set only. Results demonstrate the effectiveness of the proposed relation modules. Both R-GCN and HRCA make positive contributions to the performance. The \textbf{w HRCA(FC)} denotes the combination version HRCA with R-GCN in Fig.~\ref{fig:hrca_agg}. The \textbf{w HRCA(C2F)} denotes the coarse-to-fine version HRCA. Note that all ablations are conducted with CNN-LSTM-based MURPHY.}
		\label{tab:ablation_results}
		\begin{tabular}{  l | rr | rrr | rrr   }
			\hline
			\multirow{2}{*}{{\textit{RLLS}}}
			&\multicolumn{2}{c}{{}}   
			&\multicolumn{3}{c}{{Average precision}}  
			&\multicolumn{3}{c}{{}}      \\ 
			\cline{2-9}

			& \textbf{SAP6} & \textbf{SAP3} & S & T & $\langle$IAO$\rangle$ & I & A & O \\ \hline
			\textbf{w HRCA(FC)} & 65.27 & 69.86 & 82.48 & 68.59 & 58.52 & 60.84  & 57.45 & 63.72\\
			\textbf{w HRCA(C2F)} & 66.52 & 70.47 & 85.95 & 67.63 & 57.83 & 62.60  & 60.10 & 64.98\\
			\textbf{w R-GCN} & 67.78 & 72.04 & 87.36 & 68.88 & 59.87 & 64.25 & 61.64 & 64.65\\

			    \hline
       		MURPHY(CNN) & 69.24 & 73.10 & 87.76 & 70.57 & 60.96 & 65.58 & 63.40 & 67.18 \\
			MURPHY(CNN-LSTM) & 68.53 & 73.16 & 88.01 & 70.18 & 61.30 & 64.22 & 61.15 & 66.33\\
			\hline
		\end{tabular}
	\end{center}
\end{table}

Results of HRCA manifest the importance of inter-relational constraints from the surgical domain knowledge. We observe that the componenet-embedding-based HRCA for all inter-relations cannot sufficiently improve the performance, but this degradation can be eliminated by introducing the coarse-to-fine relation constraint. In addition, the HRCA scheme favors the relation-fused feature better than using the backbone feature alone, indicating that the relation module produces more distinguishable information. These observations suggest that the relations from domain knowledge can facilitate training convergence, demonstrating the effectiveness of our inter-relation learning module. 
Lastly, results show that the integration of the proposed relation module achieves the most significant improvement on performance among all configurations. This manifests that the intra- and inter-relation learning modules can jointly facilitate the model performance by incorporating useful relational information into a feature representation.

\subsection{Qualitative Results}
We present prediction results of a 16-minute challenging RLLS case in Fig.~\ref{fig:qualitative_results2}. This sequence is a challenging case that contains effective periods interrupted by under-effective fragments. Results show that our MURPHY achieves the best edit distance as well as satisfactory prediction results on the 3 primary tasks. Moreover, MURPHY realizes joint improvements in edit distance and prediction accuracy compared with the CNN-LSTM baseline, suggesting that the relation module is able to produce satisfactory results on the sequence with complex temporal order. We notice that all mentioned methods are prone to generating segments of under-effective frames in their predictions, especially in the stages that require complex instrument manipulations. This is a real-world problem because the expert surgeon cannot guarantee that all their movements are necessary. For instance, the surgeon waits until the completion of the assistant's operation or adjusts the instrument pose before executing an activity. These behaviors produce confusing view content that contains the same instrument and objects to the effective frames, resulting in a challenging recognition compared with the existing ideal datasets. Although the scenarios in our dataset are complicated, results show that our relation-based method achieves a satisfactory result with a temporal tolerance, indicating the relational information has the potential to provide discriminative information to address the identification of under-effective frames in surgical workflow analysis.

\begin{figure}[]
\centerline{
\includegraphics[width=1.0\linewidth]{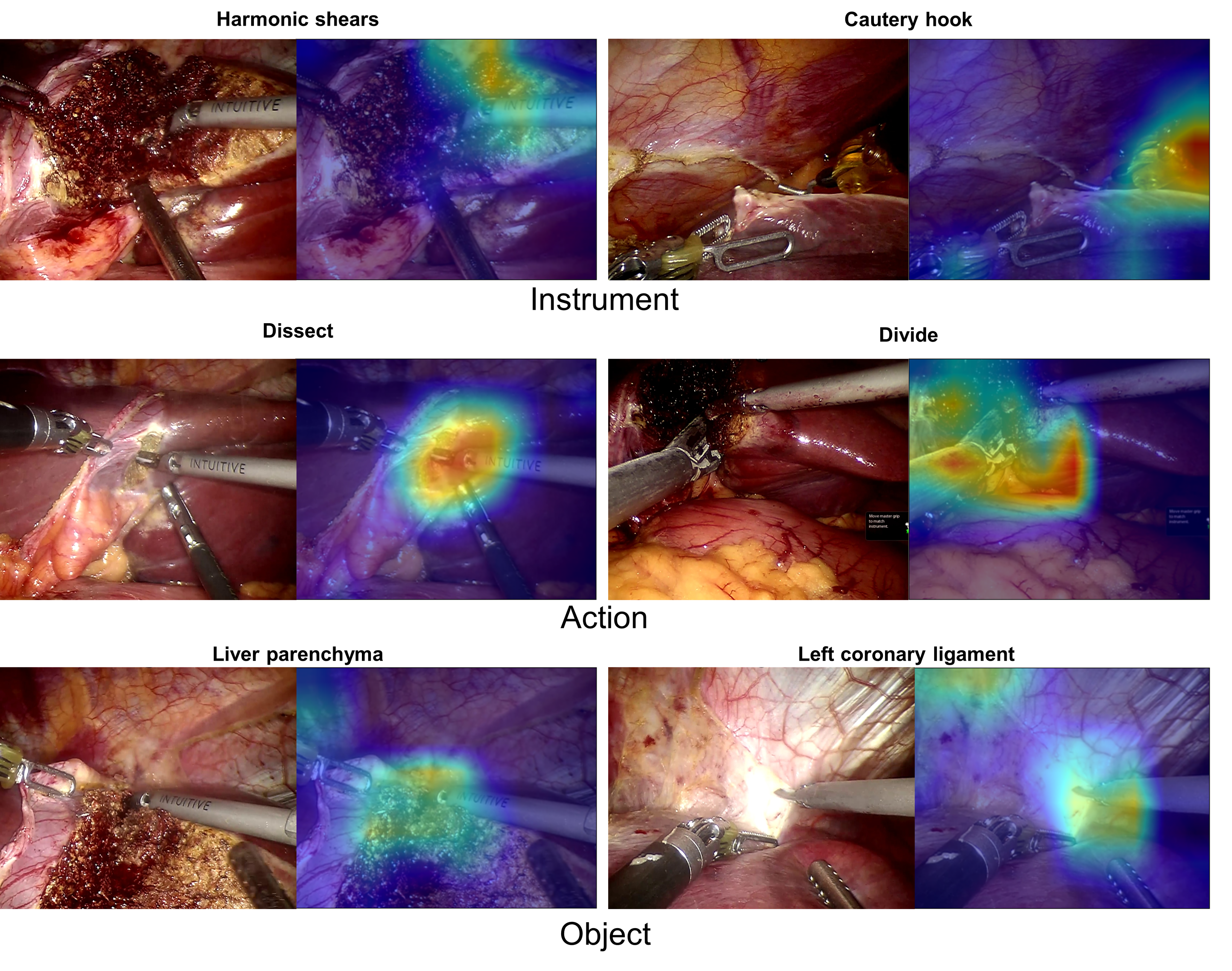}
}
\caption{Grad-CAM visualizations for instrument, action, and object. The results are from the last convolutional layer in the MURPHY's backbone module. }
\label{fig:cam_results}
\end{figure}

To verify how the model understands the triplet activity, we conduct the Grad-CAM~\cite{selvaraju2017gradcam} visualization for triplet component localization, as shown in Fig.~\ref{fig:cam_results}. Results show that our method can identify the main instrument but may occasionally split its attention on other instruments when multiple ones exist. This is potential because a particular activity always requires a fixed combination of instruments, losing the distinguishable information among instruments. MURPHY can identify the regions of instruments without a clear indication of spatial locations from the abstract instrument labels, indicating the potential to provide a rough indication for spatial component localization. Regions of objects are generally located at the tissue near the operating instrument or the discriminative tissues, {\it e.g.}, liver parenchyma, suggesting that the localization provides some levels of interpretation on the model's inference. The localization of action is attractive in general cases, which surprisingly covers the regions that are centered around the active instruments. Another observation is that it usually identifies a broad region that partially covers both the instrument and the targeted tissues. In summary, these results show that the relational feature does not harm the localization of activity components.

\section{Discussion}
\textbf{Relations matter in surgical workflow analysis but still needs long-term research.} Surgical workflow analysis plays a significant and essential role in autonomous robotic surgery but existing works have poor investigations on relation-based methods. We report a systematical investigation of different types of relations with MURPHY. One significant finding is that hierarchical annotation facilitates model performance by offering comprehensive information from different interpretation granularities. The supervision of hierarchical annotations has been confirmed from the experimental results, implying that the domain knowledge facilitates the network's representation power. Another important finding is that the inter- and intra- relations can provide a different perspective for neural networks to comprehend the complex RLLS procedure, which has illustrated from the proposed framework. We show that the relation modules can have some levels of maneuverability by integrating domain knowledge of relations. Although we demonstrate superior performance on mAP, there is a necessity to explore a better mechanism to integrate temporal consistency with relation modeling to improve the performance on the classification and the order prediction. We show that the component information as supervision can regularize the latent feature for activity recognition, benefiting from the supervision of all auxiliary task classifiers. The localization of components further strengthens the insight. These observations manifest that multiple annotations provide effective constraints with networks for understanding complex procedure. Explorations from the proposed relation modules may help the community to investigate procedure-specific workflow analysis algorithms by benefiting from relational information in the domain knowledge.

\textbf{The relational feature shows some levels of robustness in surgical styles based on our split configuration.} The split configuration of our RLLS12M dataset is in a surgeon-wise manner that provides a testbed to examine the capability of understanding varying expertise styles. Table \ref{tab:surgeon_results} manifests that the operation styles of the 3 experts in the testing set are different. Specifically, surgeon E prefers different instruments and is less experienced compared with surgeons A and D, but the relation-based MURPHY still achieves satisfactory results compared with other methods. This observation manifests that the relational information is robust to the operating styles of surgeons, which suggests that MURPHY purifies the relational features without significantly suffering interference from the variances of data patterns. In addition, the results of \textit{RLLS-w-RC} further confirm that MURPHY has better knowledge-transferring ability that benefits from the proposed relation modules. 

\textbf{There is a performance gap between human and AI assessments in the RLLS12M dataset, but we also observe inspiring indications.} To compare AI annotations with human results, we invite 3 resident surgeons to annotate 3 very challenging surgery cases in the \textit{RLLS} testing set. The corresponding mAP for step, task, and activity is 84.57, 65.16, and 66.80. Both results are superior to our method. The human activity mAP outperforms ours by at least 20.0. But MURPHY achieves 69.68 and 59.31 on step and task recognition, which shows that the AI method has the potential to reach a comparable prediction quality to surgeons with some levels of expertise. In this paper, we investigate the incorporation of relationship as a main cue to boost the algorithm performance and achieve some success along this line. In future, better algorithms for modeling relationship can be devised. In addition, surgical videos present another important cue in the scene, that is, the shape of entities such as liver, lesion, surgical instruments, etc. We plan to investigate this shape cue \cite{barbu2010automatic,zhou2010shape,park2019deepsdf} in future.

\textbf{Our RLLS12M dataset contains abundant real-world scenarios, providing a new incubator for developing AI algorithms closer to the clinical environment.} All experimental results on mAP are mostly below 90.0, especially for activity recognition, because our RLLS12M dataset contains abundant under-effective frames and real-world challenging signals such as smoke, specularity, and blood leaking. Consequently, our dataset is relatively challenging but closer to the clinical needs due to the real-world noisy signals which suppose to require more precise efforts, such as specularity and smoke removal, to address those issues. It is reasonable to be challenging for existing methods due to the lack of real-world scenarios that happened during the deployment. Therefore, our dataset has the potential to incubate powerful algorithms closer to the clinical settings, reducing the gaps between theoretical research and pressing clinical needs. We will make the RLLS12M more versatile by incorporating more clinical environment-oriented tasks and more data samples in the future. It is also valuable to explore the identification of types of under-effective frames and their correlation with the surgeon's skill level, which will be the tasks in our future work.

\section{Conclusion}
We propose MURPHY, a new hybrid relation learning framework that augments the feature representation with useful relational information for hierarchical surgical workflow analysis. The relation modules in MURPHY utilize relational graph convolution and cross-attention mechanisms to learn from intra- and inter-relations. We contribute the first RLLS dataset, RLLS12M, with hierarchical workflow annotations, including abundant annotation entities and relations, offering the opportunity to apply relations in analyzing the surgical workflow. Experimental results demonstrate the effectiveness of our approach, manifesting that relations matter in surgical workflow analysis.

\ifCLASSOPTIONcaptionsoff
  \newpage
\fi



\bibliographystyle{IEEEtran}
\bibliography{abare_jrnl_compsoc}
\end{document}